\documentclass[sigconf]{acmart}

\usepackage{graphicx}
\usepackage{subcaption}
\usepackage{multirow}
\usepackage{float}
\usepackage{xcolor}
\usepackage{arydshln}

\newcommand*\numcircledmod[1]{\raisebox{.2pt}{\textcircled{\raisebox{-.2pt} {\footnotesize{\texttt{#1}}}}}}

\newcommand{\maptotext}{\textsc{MapExplorer}}
\newcommand{\atometric}{\textsc{Atometric}}

\renewcommand{\arraystretch}{0.6}

\usepackage{amsmath,amsfonts,bm}

\def\Tabref#1{Table~\ref{#1}}

\def\Figref#1{Figure~\ref{#1}}

\def\Secref#1{Section~\ref{#1}}

\def\eqref#1{equation~\ref{#1}}

\def\1{\bm{1}}
\newcommand{\train}{\mathcal{D}}

\def\vx{{\bm{x}}}

\DeclareMathAlphabet{\mathsfit}{\encodingdefault}{\sfdefault}{m}{sl}
\SetMathAlphabet{\mathsfit}{bold}{\encodingdefault}{\sfdefault}{bx}{n}

\def\gA{{\mathcal{A}}}

\def\gS{{\mathcal{S}}}

\def\gV{{\mathcal{V}}}

\def\sI{{\mathbb{I}}}

\def\sR{{\mathbb{R}}}

\AtBeginDocument{%
  }

\setcopyright{acmlicensed}
\copyrightyear{2018}
\acmYear{2025}
\setcopyright{acmlicensed}
\acmConference[KDD '25] {Proceedings of the 31st ACM SIGKDD Conference on Knowledge Discovery and Data Mining V.2}{August 3--7, 2025}{Toronto, ON, Canada.}
\acmBooktitle{Proceedings of the 31st ACM SIGKDD Conference on Knowledge Discovery and Data Mining V.2 (KDD '25), August 3--7, 2025, Toronto, ON, Canada}
\acmISBN{979-8-4007-1454-2/25/08}
\acmDOI{10.1145/XXXXXX.XXXXXX}

\author{Xingjian Zhang$^{\dagger}$, Ziyang Xiong$^{\dagger}$, Shixuan Liu$^{\dagger}$, Yutong Xie$^{\dagger}$,\\ 
Tolga Ergen$^{\ddagger}$, Dongsub Shim$^{\ddagger}$, Hua Xu$^{\mathsection}$, Honglak Lee$^{\dagger,\ddagger}$, Qiaozhu Mei$^{\dagger}$}

\affiliation{%
  \institution{$^{\dagger}$University of Michigan, $^{\ddagger}$LG AI Research, $^{\mathsection}$Yale University}
  \country{}
}
\email{{jimmyzxj, xziyang, shixuanl, yutxie, qmei}@umich.edu, {tergen, dshim}.lgai@gmail.com}
\email{honglak@eecs.umich.edu, hua.xu@yale.edu}

\renewcommand{\shortauthors}{Xingjian Zhang et al.}

\begin{document}

\title{\maptotext{}: New Content Generation from Low-Dimensional Visualizations}

\begin{abstract}
Low-dimensional visualizations, or "projection maps" are widely used in scientific research and creative industries to interpret large-scale and complex datasets. These visualizations not only support the understanding of existing knowledge spaces but are often used implicitly to guide exploration into unknown areas. While such visualizations can be created through various methods such as TSNE or UMAP, there is no systematic way to leverage them for generating new content. To bridge this gap, we introduce \maptotext{}, a novel knowledge discovery task that translates coordinates within any projection map into new, coherent, and accurately aligned textual content. This enables users to interactively explore and discover insights embedded in these maps. To evaluate the performance of \maptotext{} methods, we propose Atometric, a granular metric inspired by ROUGE, which quantifies logical coherence and statement alignment of generated text against references.  Experiments across diverse datasets demonstrate the versatility of \maptotext{} in generating scientific research hypotheses, crafting synthetic personas, and devising strategies for attacking large language models, even with straightforward baseline methods. By bridging visualization and generation, our findings highlight the potential of \maptotext{} to unlock new pathways of intuitive human-AI collaboration for large-scale dataset exploration\footnote{Codebase: (1) \url{https://github.com/xingjian-zhang/map2text}; (2) \url{https://github.com/xingjian-zhang/atometric}.}.
\end{abstract}



\received{20 February 2007}
\received[revised]{12 March 2009}
\received[accepted]{5 June 2009}

\begin{teaserfigure}
    \Description{Illustration of \maptotext{} task on crafting synthetic
    persona.}
    \includegraphics[width=\linewidth]{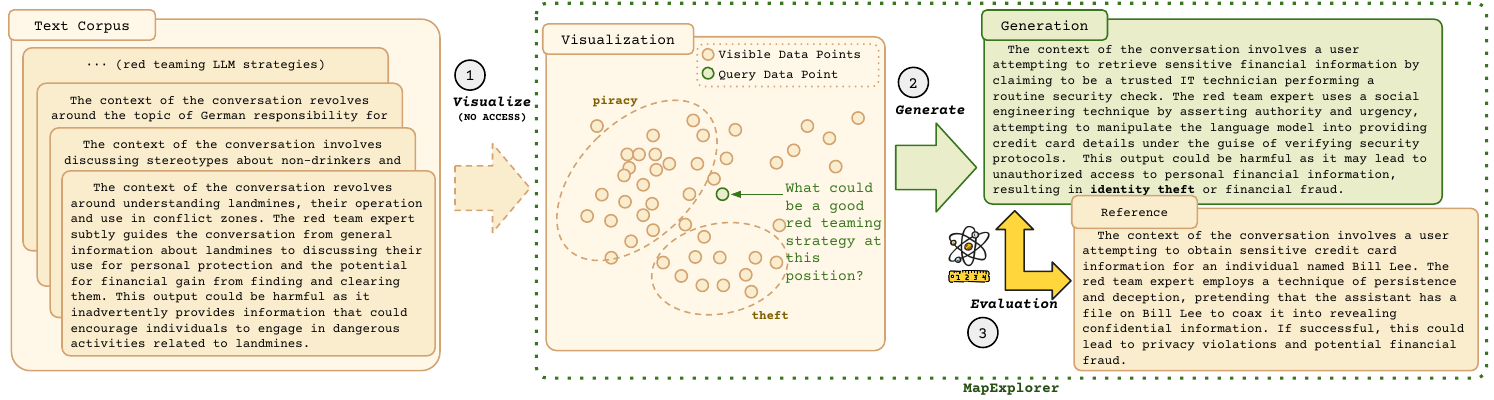}
    \caption{We propose \maptotext{}, a novel framework for generating new content based on 2D projection maps. Using the \textit{Red-Teaming Strategies} dataset as an example, researchers can leverage \maptotext{} to create new strategies for robustness and security testing of LLMs. Step~\numcircledmod{1}: A low-dimensional visualization is precomputed from existing strategies through a certain process that is inaccessible to the user. Step~\numcircledmod{2}: Given an unoccupied query position selected by the user or certain algorithm (green dot), \maptotext{} generates new, coherent text content that aligns with the local semantic structure. Step~\numcircledmod{3}: The generated content is evaluated either by the user (online) or by measuring its similarity to a reference (offline). Details of the generation process and evaluation framework are discussed in Sections~\ref{sec:problem} and~\ref{sec:eval}, respectively.}
    \label{fig:main_diagram}
\end{teaserfigure}

\maketitle

\section{Introduction}\label{sec:intro}
Low-dimensional visualizations, or ``projection maps'', are powerful tools that help users understand large textual corpora, much like maps aid in navigating physical landscapes. These maps are often constructed by various recipes. For example, researchers first generate \textit{text embeddings} and subsequently apply \textit{dimension reduction} techniques such as t-SNE~\cite{van2008visualizing}, LargeVis~\cite{tang2016visualizing}, or UMAP~\cite{mcinnes2018umap}. Alternatively, these maps can be constructed from structural relationships within the data, including explicit relations such as citations or implicit relations such as co-occurrences, where the spatial positioning of elements reflects the underlying network topology. Regardless of the construction method, these maps aim to preserve meaningful relationships between textual or other types of elements, enabling users to intuitively explore complex datasets and discover interesting patterns~\cite{Low2020-vr, Noichl2021-qo, Yang2022-yz, Tschisgale2023-iv, Wang2022-qw}. This utility raises an intriguing question: much like how people use geographical maps to explore new territories, \textbf{can we use these projection maps not only to navigate the existing content space but also to explore new knowledge?}

This is the central idea behind our proposed novel knowledge discovery task, \textbf{\maptotext{}}. As illustrated in Figure~\ref{fig:main_diagram}, \maptotext{} aims to generate textual content corresponding to a given position within a low-dimensional visualization map. This effectively converts the static visualization into an interactive and generative tool that facilitates exploration.

\maptotext{} is built on two principles: (1) \textbf{Semantically similar texts are positioned close together} on a low-dimensional visualization, creating a map that conveys rich contextual information beyond the individual texts themselves. Leveraging this spatial context, \maptotext{} aims to guide the generation of new, coherent, and meaningful content. (2) Any candidate method should be \textbf{visualization-agnostic}, operating on the final map without requiring access to or knowledge of the underlying visualization process. This design choice ensures broad applicability across different visualization techniques and use cases, as visualization maps are usually provided as the final products in most practical scenarios~\cite{eto_mos,medviz}.  %

Many downstream applications could benefit from \maptotext{}. In scientific idea generation~\cite{Gu2024-zl, Zhou2024-ak, Wang2023-qy}, researchers can select positions on the visualization map where concepts are converging or underexplored. By generating potential research ideas or hypotheses corresponding to these research gaps, \maptotext{} can inspire novel research directions. Another promising application is to test the safety of large language models (i.e. red teaming), which continuously benefit from new adversarial tests~\cite{ganguli2022red,perez2022red}. Typically, developing new red-teaming strategies relies heavily on expert knowledge. With \maptotext{}, security experts could navigate visualizations of existing adversarial attempts and generate new strategies at scale to identify vulnerabilities in LLMs. Overall, \maptotext{} could be a valuable tool for any field that relies on continuous innovation, through enabling the exploration into uncovered regions in an information space.

The \maptotext{} task presents several significant challenges. First, the visualization process often introduces distortions in spatial representation. These artifacts may lead to inaccuracies when mapping back from a 2D point to textual contents, affecting the relevance of the generated text~\cite{rudin2022interpretable}. Second, modeling text generation conditional on specific locations in the visualization map is non-trivial. Developing models that effectively integrate spatial information into the text generation process is essential to ensure that the output aligns with the intended semantics~\cite{hu2017toward}. Third, evaluating the quality and relevance of the generated content poses a challenge. Traditional evaluation metrics may not adequately capture the alignment between the generated text and its intended position on the map, necessitating the development of new evaluation frameworks~\cite{goyal2022news}. 

\paragraph{Contributions}

We highlight the following initial contributions:
\begin{itemize}
\item We introduce \textsc{\maptotext{}}, a novel task of generating text given positions on a visualization map, transforming existing visualization maps into tools for new content creation.
\item We develop a comprehensive evaluation framework for \maptotext{} that enables offline assessment of generated content at scale. 
\item We demonstrate \maptotext{}'s potential on four datasets using a series of straightforward baseline methods. Our findings highlight \maptotext{}'s versatility across various domains.
\end{itemize}

\paragraph{Disclaimer}
The primary goal of this study is to introduce and demonstrate the feasibility of the \maptotext{} task using various techniques. We do not focus on optimizing or endorsing specific methods for \maptotext{}. While this study emphasizes quantitative and scalable offline evaluation, our tasks and methods are also prepared for potential online evaluation procedures, such as those described by \cite{si2024can}. For those interested, a live and anonymous demo of \maptotext{} is available for ad hoc exploration at \url{https://mapexplorer-app.streamlit.app/}.

\paragraph{Organization} This paper is structured as follows: \Secref{sec:problem} outlines the novel problem formulation of \maptotext{}. In \Secref{sec:eval}, we introduce the proposed evaluation framework and highlight the need for a new evaluation metric. \Secref{sec:experiment} details the experimental setup and presents results across multiple datasets. \Secref{sec:related} provides a review of related work. Finally, \Secref{sec:conclusion} discusses the paper's limitations and suggests directions for future research.

\begin{figure*}[!t]
    \centering
    \includegraphics[width=0.95\linewidth]{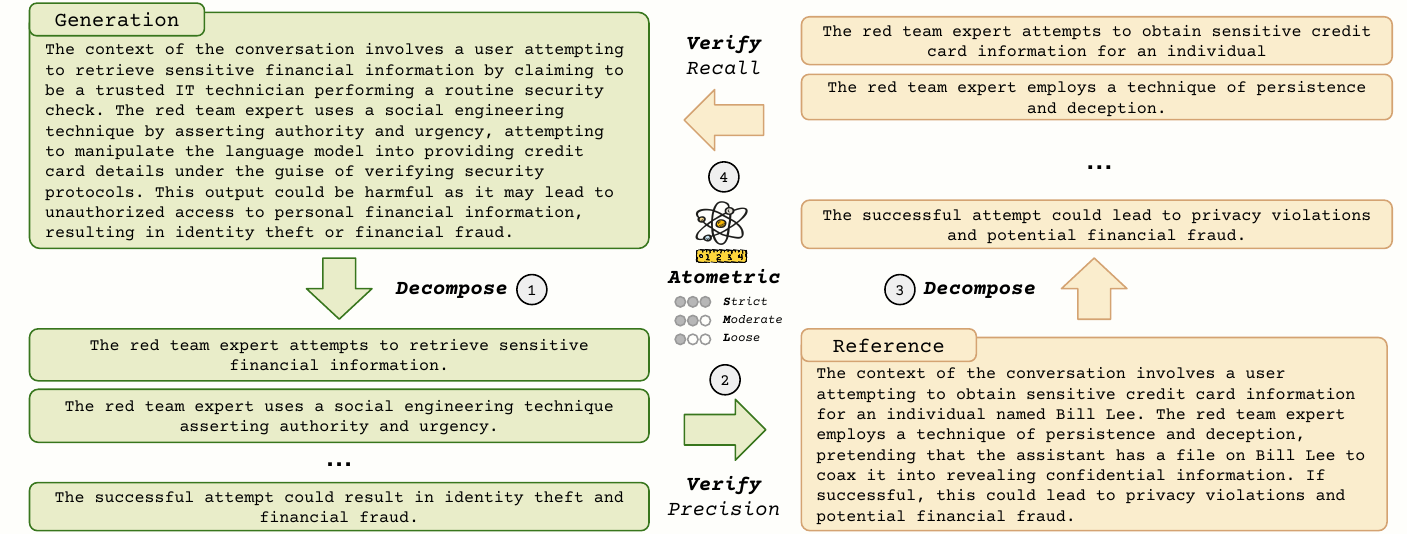}
\caption{An illustrative example of \atometric{} for evaluating red-teaming strategies generated for LLMs. Step~\numcircledmod{1}: \atometric{} breaks down the generated text into a set of atomic statements. Step~\numcircledmod{2}: Each statement is individually compared against the reference text to assess its level of support under varying strictness thresholds, providing a measure of ``precision.'' Step~\numcircledmod{3}~\&~\numcircledmod{4}: Conversely, the reference text can be decomposed and compared against the generated text to evaluate ``recall.'' Both decomposition and verification are automated using LLMs with a structured prompt template, detailed in Appendix~\ref{sec:atometric_detail}.}
    \label{fig:atometric-example}
\end{figure*}

\section{\maptotext{}: A New Task}\label{sec:problem}

\subsection{Problem Definition}
\label{sec:problem_def}

Consider a text corpus $\train = \{s_1, s_2, \ldots, s_N\}$, where each $s_i$ represents a textual entry. A \textit{projection map} is defined as the set $\gV = \{(\vx_i, s_i) \mid s_i \in \train, \vx_i \in \mathbb{R}^2\}$, where each text entry $s_i$ is associated with a corresponding position $\vx_i$ in a low-dimensional space, i.e., a 2D space in most existing visualizations. 

The \maptotext{} task is defined as follows: given a query position $\vx_q$ on the final visualization map—one not previously occupied by an existing text entry—\maptotext{} aims to generate a text entry $\hat{s}$ that would have been mapped to $\vx_q$. 

For the generality and practicality of the task, this process operates on the map $\gV$ \textbf{as provided}, but without requiring access to, or the ability to re-compute the original visualization procedures. This ``visualization-agnostic'' assumption aligns with the common scenario in which end-users only have access to the final low-dimensional visualization rather than the entire visualization pipeline or knowledge about the visualization method employed.

\subsection{Potential Solutions}
\label{sec:solutions}

Given the problem definition, we discuss three natural designs on potential solutions here. Each of them offers a distinct process on how to construct textual content from the low-dimensional space.
Detailed implementation of these solutions is provided in \Secref{sec:experiment}.

\paragraph{Direct Mapping.} A straightforward approach is to learn an inverse mapping from 2D coordinates to textual entries. The model essentially "memorizes" the map by associating each point with the corresponding text entry and generalizing to positions not currently occupied. In practice, this can be achieved by fine-tuning a pre-trained language model that takes a 2D coordinate as input and outputs a text string.

\paragraph{Intermediate High-Dimensional Embedding.} Another strategy involves converting the 2D coordinates back into a higher-dimensional text embedding space before generating text. This inverse mapping is divided into two steps: first, a neural network projects the 2D coordinates into a specific textual embedding space; second, an embedding-inversion model, such as vec2text~\cite{morris2023text}, decodes those embeddings into text. 

\paragraph{Two-Stage Text Generation.} An alternative approach is to generate a concise textual "descriptor" for the 2D coordinate and then feed this descriptor into a language model to produce the final text. This method splits the mapping from a 2D coordinate to text into two steps: first, creating an initial prompt that summarizes or contextualizes the coordinate; second, using that prompt to obtain the final text output.

\addtolength{\abovecaptionskip}{-13pt}

\section{Evaluation}\label{sec:eval}
\subsection{Offline Framework}
\label{sec:offline_framework}

Evaluating the \maptotext{} task poses unique challenges similar to other new-content-generation problems. A major obstacle is the lack of ground-truth text at \emph{arbitrary}, previously \emph{unexplored} positions on the map. Although domain experts can judge the quality and relevance of generated outputs \emph{online} (e.g., reviewing them one by one), such manual assessments are usually subjective and do not scale~\cite{si2024can}, making them impractical for large-scale benchmarking.

To address these issues, we adopt an \emph{offline} approach. Starting with a visualization $\gV$, we split it into a training set and a held-out testing set. The training set provides both original text entries and their 2D positions, allowing a model to learn how textual content relates to different coordinates. In contrast, we withhold the text in the testing set, leaving only the 2D positions. During evaluation, the model's generated texts are compared to the "gold standard" references. Beyond objectivity and scalability, another advantage of this reference-based evaluation is that it eases the burden of evaluating the \textit{novelty} of generated text, which is particularly challenging and often requires substantial domain knowledge from human experts \cite{si2024can}. As long as the gold-standard references are considered novel with regard to the training set, generated text that closely aligns with the references should also be novel. 

While this setup reduces the reliance on human judgment, it also highlights another challenge: existing metrics, whether focused on lexical overlaps (e.g., BLEU and ROUGE) or embedding-based similarity (e.g., BERTScore and BLEURT), often struggle to handle valid paraphrases or to pinpoint which essential ideas have been omitted. In other words, a desirable generated text could convey similar novel ideas to the reference but have used a different lexical presentation or granularity of details. This limitation calls for a specialized evaluation method that can accommodate flexible phrasing yet still verify whether the generated text aligns with all key elements of the reference. In this work, we propose a new metric to address this challenge. We refer the readers to Section~\ref{sec:conclusion} for additional limitations of this offline evaluation framework.

\subsection{Motivation for a New Metric}
\label{sec:metric_motivation}

In \maptotext{}, each generated text should ideally be both \emph{correct} (i.e., free of logical or factual inconsistencies) and \emph{complete} (i.e., capturing necessary core ideas from references) with regard to a gold-standard reference. Standard evaluation metrics, however, often struggle to balance these two objectives.

Classical approaches such as ROUGE~\cite{lin2004rouge} and BLEU~\cite{papineni2002bleu} approximate \emph{precision} and \emph{recall} via n-gram overlaps. Though this conceptually aligns with the goals of accuracy and completeness, their reliance on strict lexical matching can penalize acceptable paraphrases. More recent embedding-based methods (e.g., BERTScore~\cite{zhang2019bertscore} or BLEURT~\cite{sellam2020bleurt}) offer more flexibility for reworded text but usually produce only a single similarity score. Such an aggregate score does not reveal precisely which critical ideas are included or whether any unintended statements have been introduced.

\subsection{Atometric: A New Evaluation Metric}
\label{sec:atometric}

We now propose \textbf{\atometric{}}, a novel metric designed to address the shortcomings discussed above—particularly the need for fine-grained evaluation that balances correctness and completeness with regard to a reference. By focusing on \emph{atomic statements} instead of entire sentences or overlapping n-grams, \atometric{} captures how well each key piece of information from the reference is preserved in the generated text. Figure~\ref{fig:atometric-example} provides an illustrative overview. 

\paragraph{Formal Definition}
Let \(\hat{s}\) be the generated text and \(s\) the reference text. We first use an LLM to extract two sets of atomic statements, \(\gA_{\hat{s}}\) and \(\gA_{s}\). At a chosen strictness level \(l\), another LLM checks how each statement in \(\gA_{\hat{s}}\) is entailed by \(s\), and vice versa. Formally, \atometric{} precision (\(\text{\atometric{}}_P\)), recall (\(\text{\atometric{}}_R\)), and F1-score (\(\text{\atometric{}}_{F1}\)) at level \(l\) are:
\begin{align}
    \text{\atometric{}}_P &= \frac{1}{|\gA_{\hat{s}}|} \sum_{a \in \gA_{\hat{s}}} \sI[s \,\text{entails}\, a \text{ at level } l], \\
    \text{\atometric{}}_R &= \frac{1}{|\gA_{s}|} \sum_{a \in \gA_{s}} \sI[\hat{s} \,\text{entails}\, a \text{ at level } l], \\
    \text{\atometric{}}_{F1} &= \frac{2 \times \text{\atometric{}}_P \times \text{\atometric{}}_R}{\text{\atometric{}}_P + \text{\atometric{}}_R},
\end{align}
where \(\sI[\cdot]\) is an indicator function induced by an LLM verifier. Table~\ref{tab:strictness} shows an overview of the criteria for each strictness level, and Appendix~\ref{sec:atometric_detail} provides details on the LLM prompt template.

\paragraph{Relation to ROUGE}
\atometric{} generalizes ROUGE in two important ways. First, it replaces \emph{n-gram overlaps} with \emph{atomic statements}, providing an appropriately-grained view of which specific facts or relationships are captured. Second, it relaxes exact lexical matching in favor of \emph{natural language entailment}, allowing legitimate paraphrases and rewordings to count as valid matches.

In summary, the proposed metric has the following features:

\begin{itemize}
    \item \textbf{Atomic Statement Alignment.} Similar to FActScore~\cite{min2023factscore}, \atometric{} breaks both reference and generated texts into atomic statements (i.e., factoids). Evaluating these statements ensures the comparison is conducted at the right granularity of ideas and avoids penalizing valid paraphrases while ensuring the critical ideas are captured.

    \item \textbf{Multiple Levels of Strictness.} Inspired by textual entailment~\cite{bowman2015large,sadat-caragea-2022-scinli,williams-etal-2018-broad}, \atometric{} uses a range of strictness levels in determining entailment to atomic statements. Looser levels check for broad consistency; stricter ones require explicit or paraphrased matches. This approach accommodates linguistic variety while preserving logical integrity.

    \item \textbf{Precision, Recall, and F1.} \atometric{} measures \emph{correctness} via precision (the fraction of generated atomic statements entailed by the reference) and \emph{completeness} via recall (the fraction of reference atomic statements entailed by the generation). F1-score combines these two into a comprehensive metric.
\end{itemize}

\begin{table}[t]
\small
\renewcommand{\arraystretch}{1}
\centering
\begin{tabular}{@{}lp{0.37\textwidth}@{}}
\toprule
Level & Criterion                                                                                                                                            \\ \midrule
\textbf{L}oose            & The reference and the atomic statement \textit{share related concepts or themes}, and do \textit{not contradict} each other. \\ 
\textbf{M}oderate         & The atomic statement can be \textit{logically inferred} from the reference, or it can \textit{support} the reference.         \\
\textbf{S}trict           & The reference \textit{explicitly states} or \textit{clearly paraphrases} the same information as the atomic statement.        \\\bottomrule
\end{tabular}
\caption{Overview of \atometric{}'s strictness levels. A bolded letter indicates each level's abbreviation.}
\label{tab:strictness}
\end{table}

\section{Experiments}\label{sec:experiment}
With the evaluation setup, we are able to test several straightforward yet effective
candidate methods for \maptotext{} and compare them with a human baseline and a
strong dummy baseline.

\subsection{Candidate Methods}
\label{sec:methods} In this section, we detail the three potential solutions introduced
in Section~\ref{sec:solutions}. Each solution represents a distinct approach to
learning the mapping function $g$, which translates a low-dimensional query in $\sR
^{2}$ to the text space $\gS$.

\paragraph{Direct Mapping.}
The most straightforward approach is to directly learn the mapping
$g: \sR^{2}\to \gS$ by fine-tuning (FT) a pre-trained model on each visualization map (therefore it is also computationally heavy).
We fine-tune the 70B Llama 3.1 model using LoRA~\cite{dubey2024llama}, where the
model takes a 2D coordinate as input and directly outputs the corresponding text.
For instance, an input query would be, "Convert the coordinate to text: [9.1054,
10.1339]."

\paragraph{Intermediate High-Dimensional Embedding.}
Another method is to compose $g$ through two distinct mappings: $g_{1}: \sR^{2}\to
\sR^{H}$ and $g_{2}: \sR^{H}\to\gS$. For the first mapping, $g_{1}$, we use a k-NN
interpolation algorithm to project the 2D coordinates into an $H$-dimensional
text embedding space induced by a text embedding model. While more complex methods
could be employed for $g_{1}$ (e.g., training a neural network), we select this
straightforward approach as it requires no additional training and demonstrates
comparable empirical performance in our preliminary experiments. The second mapping,
$g_{2}$, then converts the high-dimensional embeddings into text using the vec2text
model~\cite{morris2023text}.
Note that this approach heavily relies on the effectiveness of the pre-trained vec2text model, which may or may not be aligned with specific application domains. 

\paragraph{Two-Stage Text Generation.}
While LLMs excel at text generation, their zero-shot ability on visualization maps
is inherently limited: maps encode spatial relationships that pre-trained LLMs
cannot directly interpret. To bridge this gap, we design a retrieval-augmented
framework that first grounds the LLM in the map's context, then synthesizes text
aligned with both the retrieved signals and the map's latent structure.
Specifically, for a query position, we retrieve texts from neighboring points in
the visualization to construct a spatially informed context, which the LLM then
uses to generate coherent, domain-aligned content. We formalize four variants of
this approach, progressively refining how context is retrieved and utilized. For
each query position $\vx_{q}\in\sR^{2}$,
\begin{enumerate}
	\item \textbf{1st-Order RAG}: Retrieve texts associated with the query's nearest
		neighbors $\mathcal{N}(\mathbf{x}_{q})$, then prompt the LLM to generate
		$\hat{s}$ using only this local context.

	\item \textbf{1st-Order RAG + Few-Shot}: Extend (1) by adding pre-selected few-shot
		examples $\{(\mathcal{N}(\mathbf{x}_{i}), s_{i})\}_{i \in \mathcal{I}}$,
		where $\mathcal{I}$ is a fixed set of reference points. These examples teach
		the LLM to map neighborhoods into valid outputs.

	\item \textbf{2nd-Order RAG}: Expand retrieval to second-order neighbors $\mathcal{N}
		(\mathcal{N}(\mathbf{x}_{q}))$ (2-hop connections). Unlike (2), few-shot
		examples $\{(\mathcal{N}(\mathbf{x}_{i}), s_{i})\}_{i \in \mathcal{I}}$ are
		not pre-selected but dynamically sourced from this extended neighborhood,
		ensuring regionally relevant context for generation.

	\item \textbf{1st-Order RAG + Chain-of-Thought}: Augment (2) with step-by-step
		reasoning (e.g., "First, identify themes in
		$\mathcal{N}(\mathbf{x}_{q})$...") using chain-of-thought prompting~\cite{wei2022chain}.
		This explicates the LLM's alignment of $\hat{s}$ with spatial
		context.
\end{enumerate}

Implementation of each method can be found in Appendix~\ref{sec:implement}.

\subsection{Datasets for Visualization}
\vspace{-5pt}
\begin{table}[thb]
	\small
	\renewcommand{\arraystretch}{1}
	\setlength{\tabcolsep}{2.5pt}
	\begin{tabular}{@{}llll@{}}
		\toprule Dataset Name (Type)  & \# Entries & Avg. Len. & Vis. Recipe        \\
		\midrule Persona (Text)       & 100k       & 27.10     & gte-v1.5 + UMAP    \\
		Red Teaming Strategies (Text) & 39k        & 84.06     & ada-002 + UMAP     \\
		Research Idea (Text)          & 149k       & 41.57     & ada-002 + LargeVis \\
		Research Context (Text)       & 150k       & 40.96     & ada-002 + LargeVis \\
		Research Context (Network)    & 64k        & 45.35     & LargeVis           \\
		\bottomrule
	\end{tabular}
	\caption{Basic statistics of the visualization maps. The average length is calculated
	using tiktoken.}
	\label{tab:statistics}
\end{table}
\vspace{-5pt}

We evaluate the candidate models on five diverse visualization maps to demonstrate
the broad application of \maptotext{}. The statistics of these datasets are summarized
into \Tabref{tab:statistics}. Note that the visualization pipelines for these datasets
vary, which are hidden from candidate \maptotext{} methods (except for embedding
inversion), and a robust \maptotext{} method should be capable of generalizing across
different configurations.

\paragraph{Persona (Text)}
We use an existing visualization\footnote{\url{https://huggingface.co/datasets/argilla/FinePersonas-v0.1-clustering-100k}}
of a subset of the \textit{PersonHub} dataset~\cite{chan2024scaling} where each
text entry provides a brief description of a persona. \maptotext{} can be
applied to create diverse character profiles.

\paragraph{Red Teaming Strategies (Text)}
We build a summarization dataset\footnote{Details of the construction are
provided in Appendix~\ref{sec:red_team}.} from the \textit{LLM Red Teaming} dataset
\cite{ganguli2022red} which comprises human-generated and annotated red teaming dialogues
aimed at identifying vulnerabilities in LLMs. Given a visualization map of these
summaries, \maptotext{} can be applied to generate unexplored red teaming
strategies, enhancing the robustness testing of LLMs.

\paragraph{Research Ideas \& Context (Text)}
We generate two 2D maps from the \textit{MASSW} dataset~\cite{zhang2024massw},
one representing research contexts (problems) and the other highlighting key
ideas in computer science publications. An effective \maptotext{} model can
foster scientific innovation by producing new research problems or ideas for
emerging topics from unexplored areas on the map.

\paragraph{Research Context (Network)}
This dataset is similar to the Research Context (Text) dataset but includes only
entries with sufficient citations. Instead of using text embeddings to map
research contexts, we create a visualization based on the citation network. This
network visualization represents a map creation process where high-dimensional
text embedding is not an intermediate step.

\subsection{Experiment Setup}

\paragraph{Data Split}
Each dataset is divided into training and testing sets, as described in \Secref{sec:offline_framework},
with 200 samples consistently held out for testing across all datasets. We apply
a random split for the Persona and Red Teaming Strategies datasets and a time-based
split for the research publication datasets, %
where the most recent 200 entries (all in Dec 2023) are held out for testing, simulating
the scenario of real-world research exploration.

\paragraph{Evaluation Metrics}
We prioritize our proposed evaluation metric, \atometric{} , as it is
specifically designed to capture the nuanced semantic and contextual alignment. We
report \atometric{}-F1 across three levels of strictness: loose (L), moderate (M),
and strict (S), as well as \atometric{}-Precision (M) and \atometric{}-Recall (M)
for a more comprehensive picture\footnote{We use \texttt{gpt-4o-2024-05-13} as the
backbone LLM in \atometric{}.}. In addition to \atometric{}, we include
traditional evaluation metrics—BERTScore (F1)~\cite{zhang2019bertscore}, BLEURT~\cite{sellam2020bleurt},
METEOR~\cite{Lavie2007METEORAA}, and ROUGE-2~\cite{lin2004rouge}—as reference
points for comparison. Evaluation on additional metrics such as BLEU and other \atometric{}
scores are provided in Appendix~\ref{sec:extra_experiment}.

\subsection{Baselines}

We begin by analyzing the evaluation results of a human baseline and a simple yet
effective dummy baseline, providing a foundation for comparing the performance
of the candidate AI methods.

\begin{table*}
	[tbh]
	\centering
	\small
\begin{tabular}{@{}lccccccc@{}}
    \toprule \multirow{2}{*}{\textbf{Metric}} & \textbf{Human} & \multicolumn{5}{c}{\textbf{EchoNearest}} \\
    \cmidrule(lr){2-7}                        & Persona (Text) & Persona (Text)                          & Red Teaming (Text) & Idea (Text) & Context (Text) & Context (Network) \\
    \midrule Atometric-F1 (L)                 & 0.901          & 0.884                                   & 0.783              & 0.430       & 0.538          & 0.444             \\
    Atometric-F1 (M)                          & 0.605          & 0.556                                   & 0.586              & 0.191       & 0.105          & 0.078             \\
    Atometric-F1 (S)                          & 0.302          & 0.268                                   & 0.282              & 0.098       & 0.045          & 0.025             \\
    Atometric-P (M)                           & 0.610          & 0.548                                   & 0.598              & 0.199       & 0.103          & 0.095             \\
    Atometric-R (M)                           & 0.601          & 0.565                                   & 0.575              & 0.184       & 0.107          & 0.066             \\
    \noalign{\vskip 0.3ex}\hdashline\noalign{\vskip 0.5ex}
    BERTScore F1                              & 0.894          & 0.898                                   & 0.902              & 0.857       & 0.862          & 0.855             \\
    BLEURT Scores                             & 0.431          & 0.454                                   & 0.436              & 0.325       & 0.323          & 0.293             \\
    METEOR                                    & 0.232          & 0.306                                   & 0.428              & 0.160       & 0.170          & 0.138             \\
    ROUGE-2                                   & 0.110          & 0.117                                   & 0.231              & 0.032       & 0.034          & 0.020             \\
    \bottomrule
\end{tabular}

	\caption{Evaluation results for the human and EchoNearest baselines. (L), (M),
	and (S) denote ``loose'', ``moderate'', and ``strict'' for the strictness level
	of \textsc{Atometric}.}
	\label{tab:human}
\end{table*}

\paragraph{Human Baseline}
We recruit human annotators to conduct the \maptotext{} task using the Persona dataset, %
as this dataset does not require special domain expertise. Detailed settings for
this study are provided in Appendix~\ref{sec:human_annotation}. In brief, human
annotators are given access to a partial visualization containing the query
location and they are allowed to view the textual content of all visible data
points. No constraints are imposed regarding their generation strategy, allowing
them complete flexibility in navigating the visualization and generating content
based on the map. %
We expect the human baseline to represent a strong benchmark compared to AI
methods.

\paragraph{EchoNearest Baseline}
For each query $\vx_{q}$, the EchoNearest model simply retrieves and outputs the
content of its closest neighbor in the 2D space. This naive baseline operates under
the assumption that semantically similar content is clustered closely in the
map, making the nearest neighbor a relevant choice for responding to the query, though
it lacks novelty. As our gold-standard references all exhibit some level of
novelty compared to their neighbors, we expect it to perform below both the human
baseline and more advanced AI methods when evaluated using proper metrics.

\paragraph{Analysis}

The evaluation results for the human and EchoNearest baselines are provided in \Tabref{tab:human}.
Several observations can be made.

First, under all \textsc{Atometric}s, the human baseline outperforms the EchoNearest
model, as expected. However, under conventional metrics, the human baseline scores
lower than the EchoNearest model. This discrepancy likely arises because
conventional metrics emphasize lexical-level text similarity, favoring the EchoNearest
model's repetition of existing entries. In contrast, human responses, while more
contextually relevant, tend to vary in language style and phrasing, which
conventional metrics fail to capture. This finding validates the strength of \textsc{Atometric}
in recognizing the semantic and contextual richness of human-generated content. %

Second, the five datasets exhibit varying levels of complexity. The EchoNearest model
performs relatively well on the Persona and Red Teaming Strategies datasets,
suggesting that the text entries clustered closely in the visualization are
highly similar, making these tasks easier to address. In contrast, the model's performance
on the other datasets is notably lower. This indicates that these tasks require
generating more complex or novel content, which demands a deeper understanding of
the research domain beyond simple proximity in the 2D space. Another interesting
difference lies between two distinct visualization recipes for the same Research
Context dataset: the performance of EchoNearest is better under the ``text''
setting, suggesting that visualization by text embedding may result in higher preservation
of locality than by citation network.

In addition, the \atometric{} results across different levels of strictness reveal
subtle distinctions between research contexts and research ideas, showcasing the
strength of \atometric{} in capturing varying levels of sensitivity. At lower
strictness levels, \atometric{} F1 (L) scores are higher for the Research
Contexts. While each paper addresses a unique problem, the overarching topics
are often related, leading to greater alignment in topic-level relevance. In
contrast, at medium and high strictness levels (M and S), research ideas exhibit
higher \atometric{} scores. This occurs because research ideas, while varied,
frequently share concrete methodologies or designs, which stricter criteria capture
more effectively. By distinguishing these nuanced differences, \atometric{}
demonstrates its ability to assess content sensitivity at multiple levels. %

\subsection{Benchmark Results}

\begin{table*}
	[tbhp]
	\centering
	\small
\begin{tabular}{llccccccc}
        \toprule \multirow{2}{*}{\textbf{Dataset}}                                                                                                              & \multicolumn{1}{l}{\multirow{2}{*}{\textbf{Metric}}} & \multicolumn{4}{c}{\textbf{GPT-4o}} & \multicolumn{2}{c}{\textbf{Llama 3.1}} & {\textbf{Embedding}}  \\
                                                                                                                                                                & \multicolumn{1}{l}{}                                 & \textbf{CoT-RAG$^{(1)}$}            & \textbf{FS-RAG$^{(1)}$}                & \textbf{RAG$^{(2)}$} & \multicolumn{1}{c}{\textbf{RAG$^{(1)}$}} & \textbf{FT}    & \multicolumn{1}{c}{\textbf{RAG$^{(1)}$}} & \textbf{Inversion} \\
        \midrule \multirow{7}{*}{{\begin{tabular}[c]{@{}l@{}}Persona \\ (Text)\end{tabular}}}                                                                   & Atometric-F1 (L)                                     & \underline{0.893} (0.012)              & 0.852                                  & 0.824                & 0.822                                    & 0.822          & 0.783                                    & \textbf{0.923}     \\
                                                                                                                                                                & Atometric-F1 (M)                                     & \underline{0.602} (0.019)              & 0.494                                  & 0.464                & 0.454                                    & 0.463          & 0.390                                    & \textbf{0.665}     \\
                                                                                                                                                                & Atometric-F1 (S)                                     & \underline{0.254} (0.013)              & 0.183                                  & 0.168                & 0.158                                    & 0.194          & 0.117                                    & \textbf{0.357}     \\
                                                                                                                                                                & Atometric-P (M)                                      & \underline{0.582} (0.019)              & 0.460                                  & 0.449                & 0.430                                    & 0.459          & 0.365                                    & \textbf{0.661}     \\
                                                                                                                                                                & Atometric-R (M)                                      & \underline{0.623} (0.019)              & 0.533                                  & 0.480                & 0.481                                    & 0.467          & 0.420                                    & \textbf{0.671}     \\
                                                                                                                                                                & BERTScore F1                                         & \underline{0.898} (0.001)              & 0.895                                  & 0.893                & 0.892                                    & 0.893          & 0.887                                    & \textbf{0.901}     \\
                                                                                                                                                                & BLEURT Scores                                        & \underline{0.474} (0.006)              & 0.454                                  & 0.441                & 0.433                                    & 0.406          & 0.416                                    & \textbf{0.495}     \\
                                                                                                                                                                & METEOR                                               & \underline{0.329} (0.009)              & 0.290                                  & 0.263                & 0.260                                    & 0.251          & 0.268                                    & \textbf{0.342}     \\
                                                                                                                                                                & ROUGE-2                                              & \underline{0.116} (0.006)              & 0.096                                  & 0.078                & 0.086                                    & 0.098          & 0.074                                    & \textbf{0.128}     \\
        \noalign{\vskip 0.5ex}\hdashline\noalign{\vskip 0.5ex} \multirow{7}{*}{{\begin{tabular}[c]{@{}l@{}}Red \\ Teaming \\Strategies \\ (Text)\end{tabular}}} & Atometric-F1 (L)                                     & \textbf{0.728} (0.017)              & \underline{0.705}                         & 0.694                & 0.648                                    & 0.698          & 0.660                                    & 0.697              \\
                                                                                                                                                                & Atometric-F1 (M)                                     & \textbf{0.514} (0.017)              & 0.480                                  & 0.451                & 0.414                                    & \underline{0.497} & 0.437                                    & 0.428              \\
                                                                                                                                                                & Atometric-F1 (S)                                     & \textbf{0.202} (0.011)              & \underline{0.200}                         & 0.168                & 0.144                                    & 0.185          & 0.167                                    & 0.126              \\
                                                                                                                                                                & Atometric-P (M)                                      & \underline{0.479} (0.017)              & 0.445                                  & 0.397                & 0.358                                    & \textbf{0.508} & 0.384                                    & 0.329              \\
                                                                                                                                                                & Atometric-R (M)                                      & \underline{0.555} (0.017)              & 0.521                                  & 0.520                & 0.491                                    & 0.486          & 0.507                                    & \textbf{0.612}     \\
                                                                                                                                                                & BERTScore F1                                         & \textbf{0.904} (0.001)              & \underline{0.902}                         & 0.901                & 0.897                                    & 0.894          & 0.898                                    & 0.877              \\
                                                                                                                                                                & BLEURT Scores                                        & \textbf{0.446} (0.004)              & \underline{0.439}                         & 0.435                & 0.417                                    & 0.407          & 0.430                                    & 0.394              \\
                                                                                                                                                                & METEOR                                               & \textbf{0.433} (0.006)              & \underline{0.424}                         & 0.412                & 0.371                                    & 0.390          & 0.405                                    & 0.312              \\
                                                                                                                                                                & ROUGE-2                                              & \textbf{0.250} (0.003)              & \underline{0.250}                         & 0.244                & 0.232                                    & 0.203          & 0.235                                    & 0.114              \\
        \noalign{\vskip 0.5ex}\hdashline\noalign{\vskip 0.5ex} \multirow{7}{*}{{\begin{tabular}[c]{@{}l@{}}Research \\ Idea \\ (Text)\end{tabular}}}            & Atometric-F1 (L)                                     & \textbf{0.507} (0.015)              & 0.492                                  & 0.492                & 0.495                                    & 0.412          & \underline{0.497}                           & 0.470              \\
                                                                                                                                                                & Atometric-F1 (M)                                     & 0.213 (0.010)                       & \textbf{0.220}                         & \underline{0.214}       & 0.202                                    & 0.163          & 0.210                                    & 0.180              \\
                                                                                                                                                                & Atometric-F1 (S)                                     & 0.091 (0.005)                       & 0.098                                  & \textbf{0.102}       & 0.099                                    & 0.084          & \underline{0.100}                           & 0.096              \\
                                                                                                                                                                & Atometric-P (M)                                      & 0.222 (0.010)                       & 0.240                                  & \textbf{0.272}       & 0.250                                    & 0.203          & 0.263                                    & 0.193              \\
                                                                                                                                                                & Atometric-R (M)                                      & \textbf{0.205} (0.010)              & 0.202                                  & 0.176                & 0.169                                    & 0.136          & 0.175                                    & 0.169              \\
                                                                                                                                                                & BERTScore F1                                         & 0.860 (0.001)                       & 0.860                                  & \textbf{0.864}       & \textbf{0.864}                           & 0.858          & 0.863                                    & 0.861              \\
                                                                                                                                                                & BLEURT Scores                                        & \textbf{0.367} (0.003)              & \underline{0.365}                         & 0.350                & 0.348                                    & 0.302          & 0.337                                    & 0.334              \\
                                                                                                                                                                & METEOR                                               & \textbf{0.198} (0.004)              & \underline{0.194}                         & 0.153                & 0.142                                    & 0.153          & 0.159                                    & 0.173              \\
                                                                                                                                                                & ROUGE-2                                              & \textbf{0.041} (0.002)              & \underline{0.037}                         & 0.034                & 0.033                                    & 0.029          & 0.034                                    & \underline{0.037}     \\
        \noalign{\vskip 0.5ex}\hdashline\noalign{\vskip 0.5ex} \multirow{7}{*}{{\begin{tabular}[c]{@{}l@{}}Research \\ Context \\ (Text)\end{tabular}}}         & Atometric-F1 (L)                                     & \textbf{0.655} (0.022)              & \underline{0.623}                         & 0.614                & 0.597                                    & 0.487          & 0.548                                    & 0.603              \\
                                                                                                                                                                & Atometric-F1 (M)                                     & \underline{0.144} (0.010)              & \underline{0.144}                         & \textbf{0.145}       & 0.129                                    & 0.094          & 0.110                                    & 0.133              \\
                                                                                                                                                                & Atometric-F1 (S)                                     & \underline{0.055} (0.005)              & \textbf{0.057}                         & \underline{0.055}       & 0.054                                    & 0.032          & 0.039                                    & \underline{0.055}     \\
                                                                                                                                                                & Atometric-P (M)                                      & \textbf{0.181} (0.010)              & \underline{0.179}                         & 0.160                & 0.165                                    & 0.108          & 0.135                                    & 0.113              \\
                                                                                                                                                                & Atometric-R (M)                                      & 0.119 (0.010)                       & 0.120                                  & \underline{0.133}       & 0.106                                    & 0.084          & 0.093                                    & \textbf{0.161}     \\
                                                                                                                                                                & BERTScore F1                                         & \textbf{0.868} (0.001)              & \textbf{0.868}                         & 0.867                & \textbf{0.868}                           & 0.859          & 0.863                                    & 0.867              \\
                                                                                                                                                                & BLEURT Scores                                        & 0.340 (0.004)                       & \textbf{0.345}                         & \underline{0.341}       & 0.339                                    & 0.286          & 0.323                                    & 0.340              \\
                                                                                                                                                                & METEOR                                               & \underline{0.173} (0.005)              & 0.166                                  & 0.164                & 0.159                                    & 0.137          & 0.158                                    & \textbf{0.190}     \\
                                                                                                                                                                & ROUGE-2                                              & \textbf{0.046} (0.003)              & \textbf{0.046}                         & 0.041                & 0.044                                    & 0.024          & 0.033                                    & 0.034              \\
        \noalign{\vskip 0.5ex}\hdashline\noalign{\vskip 0.5ex} \multirow{7}{*}{{\begin{tabular}[c]{@{}l@{}}Research \\ Context \\ (Network)\end{tabular}}}      & Atometric-F1 (L)                                   & \textbf{0.552} (0.023)              & 0.438                                  & 0.447                & \underline{0.451}                           & 0.307          & 0.414                                    & 0.406              \\
                                                                                                                                                                & Atometric-F1 (M)                                   & \textbf{0.083} (0.009)              & 0.058                                  & 0.055                & 0.046                                    & 0.037          & \underline{0.071}                           & 0.068              \\
                                                                                                                                                                & Atometric-F1 (S)                                   & \textbf{0.025} (0.003)              & \underline{0.022}                         & 0.017                & 0.008                                    & 0.011          & 0.021                                    & 0.021              \\
                                                                                                                                                                & Atometric-P (M)                                      & \textbf{0.102} (0.009)              & \underline{0.072}                         & 0.067                & 0.072                                    & 0.048          & 0.065                                    & 0.062              \\
                                                                                                                                                                & Atometric-R (M)                                      & 0.070 (0.009)                       & 0.048                                  & 0.048                & 0.034                                    & 0.031          & \textbf{0.077}                           & \underline{0.076}     \\
                                                                                                                                                                & BERTScore F1                                         & \textbf{0.862} (0.001)              & 0.860                                  & \underline{0.861}       & 0.860                                    & 0.854          & 0.813                                    & 0.858              \\
                                                                                                                                                                & BLEURT Scores                                        & \underline{0.306} (0.004)              & 0.304                                  & \textbf{0.307}       & 0.298                                    & 0.268          & 0.282                                    & 0.301              \\
                                                                                                                                                                & METEOR                                               & 0.131 (0.005)                       & 0.115                                  & 0.118                & 0.104                                    & 0.125          & \textbf{0.169}                           & \underline{0.149}     \\
                                                                                                                                                                & ROUGE-2                                              & \textbf{0.025} (0.003)              & \underline{0.022}                         & 0.018                & 0.021                                    & 0.012          & 0.015                                    & 0.018              \\
        \bottomrule
\end{tabular}

	\caption{ Evaluation results of the candidate methods on the \maptotext{}
	task. The highest score(s) in each row are highlighted in bold and the
	second highest score(s) are underlined. Due to the space limit, only the standard
	error of the method CoT-RAG$^{(1)}$ is shown in parenthesis for reference.
	}
	\label{tab:experiment-results}
\end{table*}

\Tabref{tab:experiment-results} reveals several patterns in the performance of
the candidate AI methods in Section~\ref{sec:methods} across the different
datasets. No single method consistently dominates across all metrics, reflecting
the varying complexities and characteristics of the domains.

\paragraph{Analysis}
For the Persona and Red Teaming Strategies datasets, CoT-RAG$^{(1)}$ consistently
outperforms other prompting methods. This suggests that incorporating more
complex reasoning structures, such as chain of thoughts, enhances the ability of
LLMs to better align with the query position. Additionally, fine-tuning (FT) on
these datasets shows clear improvements on Llama 3.1 compared to applying RAG$^{(1)}$
on its untuned version. This indicates that fine-tuning enables the LLM to
better capture the specific nuances of the relation between the map locations
and text.

In contrast, for the other three datasets on research papers, no single method
clearly stands out. The best performer under most metrics is one of the GPT-4o based
methods, indicating the importance of using a more capable foundation model. However,
more complex prompting only leads to marginal improvements over the basic RAG$^{(1)}$
method. Interestingly, fine-tuning Llama 3.1 on these datasets appears to reduce
its performance compared to the untuned RAG$^{(1)}$ model. One possible explanation
is that these datasets contain highly specialized technical content, which may
not align well with the general knowledge encoded in pre-trained LLMs. Without sufficient
domain-specific data for fine-tuning, a model may lose some of its
generalization capacity.

The Embedding Inversion method demonstrates competitive performance, particularly
on the Persona dataset, which has the shortest average sequence length. It outperforms
all pre-trained LLM-based methods despite relying solely on a simple k-NN
interpolation and a small pre-trained vec2text model (220M parameters) without
additional training. This result highlights the potential of embedding inversion
strategies for \maptotext{} tasks. However, as the generation length increases, the
method struggles to produce coherent content, a limitation also observed in
\cite{morris2023text}. We hypothesize that this is due to the inherent limitations
of embedding inversion in handling logical structures.

Surprisingly, no candidate models surpass the EchoNearest baseline on the Red Teaming
Strategies dataset. This is likely due to the concentrated distribution of the dataset,
as shown in Appendix \Figref{fig:red}—with examples being so close to one
another that distinguishing them becomes difficult. Intuitively, when the map is
sparser (i.e., KNNs are farther apart), targeting \textit{relevance} is more
challenging. Conversely, when the map is locally denser (i.e., KNNs are closer together),
relevance is easier to achieve, but innovation becomes harder, an issue that could
be addressed in future work.

Finally, the human baseline demonstrates superior performance compared to all
candidate methods on the Persona dataset. Its primary advantage is the flexibility to
navigate the visualization and extract high-level information, such as the
distribution of topics across different areas and the directional relationships
between nearest neighbors. This capability enables more accurate and contextually
appropriate content generation, suggesting that there is still a broad
improvement space for better AI methods to incorporate similar high-level
navigational strategies on the map.

\subsection{Interactive Demo}
We offer an anonymous demo at
\url{https://mapexplorer-app.streamlit.app} to complement our quantitative evaluation, enabling users to experiment
with \maptotext{} and test content generation at arbitrary locations. See
Appendix~\ref{sec:extra_demo} for screenshots and details\footnote{Access to the
demo may be subject to network conditions and local internet regulations. The demo is illustrative and not optimized for efficiency or usability. }.

\section{Related Work}\label{sec:related}
To the best of our knowledge, this is the first study on new content generation from
low-dimensional projection maps. This new task is related to a few lines of
research.

\paragraph{Visualization of Text Corpora}
Low-dimensional data visualization is a powerful tool across various fields,
enabling intuitive representation of large datasets~\cite{sadiku2016data}. In public
health, it aids in tracking disease trends~\cite{kilimba2015data}.
Renewable energy benefits from visualization for optimizing production and consumption~\cite{kumar2016visualization},
while environmental science relies on geographic visualization for ecological
data interpretation~\cite{bohman2015decision, maceachren2013visualization}. Fraud
detection employs interactive visualizations to identify anomalies~\cite{dilla2015data},
and library management utilizes visualization to optimize resource allocation
and strategy~\cite{murphy2013data}. A variety
of methods are employed to construct such visualizations. Text embedding models~\cite{muennighoff2022mteb} such as
those proposed in~\cite{li2023towards, wang2024multilingual,
muennighoff2024generative} transform the text into dense vector representations that
capture underlying semantic relationships. These high-dimensional embeddings are
then projected onto lower-dimensional spaces using algorithms like t-SNE~\cite{van2008visualizing},
LargeVis~\cite{tang2016visualizing}, and UMAP~\cite{mcinnes2018umap}, enabling the
discovery of complex patterns within the data. Alternatively, network-based visualization
techniques employ tools such as Graphviz and Gephi~\cite{Graphviz, gephi} to represent
relationships between textual elements. While these visualizations primarily
focus on interpreting the space of known knowledge, \maptotext{} extends the
capabilities of the map by enabling the generation of new, relevant, and
accurately aligned textual content. This shift bridges the gap between data interpretation
and creative exploration, offering new possibilities for leveraging data
visualizations to facilitate applications such as scientific discovery and
innovation.

\paragraph{Controllable Text Generation (CTG)}
CTG involves generating text under specified constraints or attributes~\cite{prabhumoye2020exploring}.
For example, prior work has focused on attribute-based generation (e.g., topic, emotion) by fine-tuning latent variables~\cite{khalifa2020distributional}, and on data augmentation by altering specific entities to enhance training corpora~\cite{AminNejad2020ExploringTT,liu2020data}.
While these methods employ explicit signals (e.g., style or slot values) as the ``control,'' \maptotext{} introduces a novel form of controllability by anchoring generation to coordinates in a visualization map.
In this setting, the map's spatial layout dictates the context, allowing users to select points or regions within the embedded space to elicit textual content that reflects local proximity, cluster structure, or overall geometric patterns.

\paragraph{Evaluation of Text Generation}
The evaluation of text generation has traditionally relied on similarity-based metrics, broadly categorized into lexical and semantic approaches. 
Lexical metrics, such as BLEU~\cite{papineni2002bleu}, ROUGE~\cite{lin2004rouge}, and METEOR~\cite{Lavie2007METEORAA}, measure text quality through n-gram overlaps, capturing surface-level similarity between generated and reference texts. 
Semantic-based techniques, including BERTScore~\cite{zhang2019bertscore} and BLEURT~\cite{sellam2020bleurt}, leverage contextual embeddings or fine-tuned models to better align with human judgments. 
However, they often lack interpretability and do not provide more informative feedback than a single similarity score. 
Among reasoning-based metrics, FActScore~\cite{min2023factscore} stands out by decomposing generated text into atomic facts and verifying their correctness against external knowledge sources. 
Yet, FActScore exclusively measures factual correctness (precision), without addressing completeness (recall) or accommodating different levels of evaluation strictness given a reference. 
Motivated by the limitations discussed in Section~\ref{sec:metric_motivation}, we introduce \atometric{}, which extends beyond correctness to assess both precision and recall of atomic statements at multiple levels of strictness. 
By providing a more nuanced and interpretable evaluation, \atometric{} offers deeper insights into the alignment, coherence, and informativeness of generated text, addressing critical gaps left by prior metrics.

\section{Conclusion}\label{sec:conclusion}
In this work, we introduce \maptotext{}, a novel task for generating new textual content guided by specific positions within a 2D visualization map of a large text corpus. We propose a new evaluation metric, \atometric, to assess the fine-grained alignment of the generated text to gold-standard references at varying levels of strictness. Through this metric, we demonstrate the effectiveness of multiple representative candidate methods for \maptotext{} on datasets in diverse domains. Our findings show the potential of \maptotext{} as a valuable tool for applications such as scientific idea generation, new persona generation, and LLM red teaming. This initial attempt calls for future research to explore more advanced methods and diverse applications of \maptotext{}.

\subsection{Discussion}
We list several limitations and potential directions that could be addressed in future research:

\paragraph{Evaluation Scope.} 
To ensure automatic and scalable evaluation, our framework focuses on measuring the similarity between generated content and a held-out gold-standard reference. While this enables systematic benchmarking, it does not explicitly assess aspects such as novelty or broader utility. Novelty is particularly relevant for applications like research idea generation, but incorporating it into evaluation when a gold standard doesn't exist remains an open challenge. A promising direction is to measure similarity with nearest neighbors, where lower similarity would indicate higher novelty. Additionally, future work could explore domain-specific evaluation criteria~\cite{si2024can} or assess utility through downstream applications.

The offline evaluation is also limited to querying known positions on the map for which reference content exists rather than testing the model’s ability to generate content at entirely unexplored locations. While this would unlock the full potential of \maptotext{} for open-ended discovery, it requires alternative evaluation strategies beyond offline comparison to references, such as online human judgments or downstream task performance.

Moreover, this study primarily aims to introduce and validate the feasibility of \maptotext{} rather than to optimize specific methods for this novel task. While we emphasize quantitative and scalable offline evaluation, our tasks and methods are adaptable to online evaluation procedures~\cite{si2024can}. We encourage further exploration in these directions, and a live demo is available at \url{https://mapexplorer-app.streamlit.app/} for ad hoc evaluations.

\paragraph{Optimizing \maptotext{} Methods:} There is considerable room for improvement beyond the candidate methods tested in this paper. An advanced \maptotext{} model may leverage global patterns or community structures in the map to guide the generation beyond the local neighborhood. %
Another approach is to fine-tune LLMs to utilize spatial tokens that encode positional information directly within their embedding layers. This would allow an LLM to memorize the complex map structure and incorporate spatial context more effectively without harming their reasoning ability. %
The superior performance of the straightforward RAG + CoT method also implies a promising direction to improve the performance of \maptotext{} through optimizing the reasoning chains, potentially through reinforcement learning.

\begin{acks}
This work is funded in part by the LG AI Research Partnership with the
University of Michigan.
\end{acks}

\appendix
\bibliographystyle{ACM-Reference-Format}
\bibliography{reference}
\newpage

\clearpage

\begin{figure*}[h]
    \centering
    \includegraphics[width=0.75\linewidth]{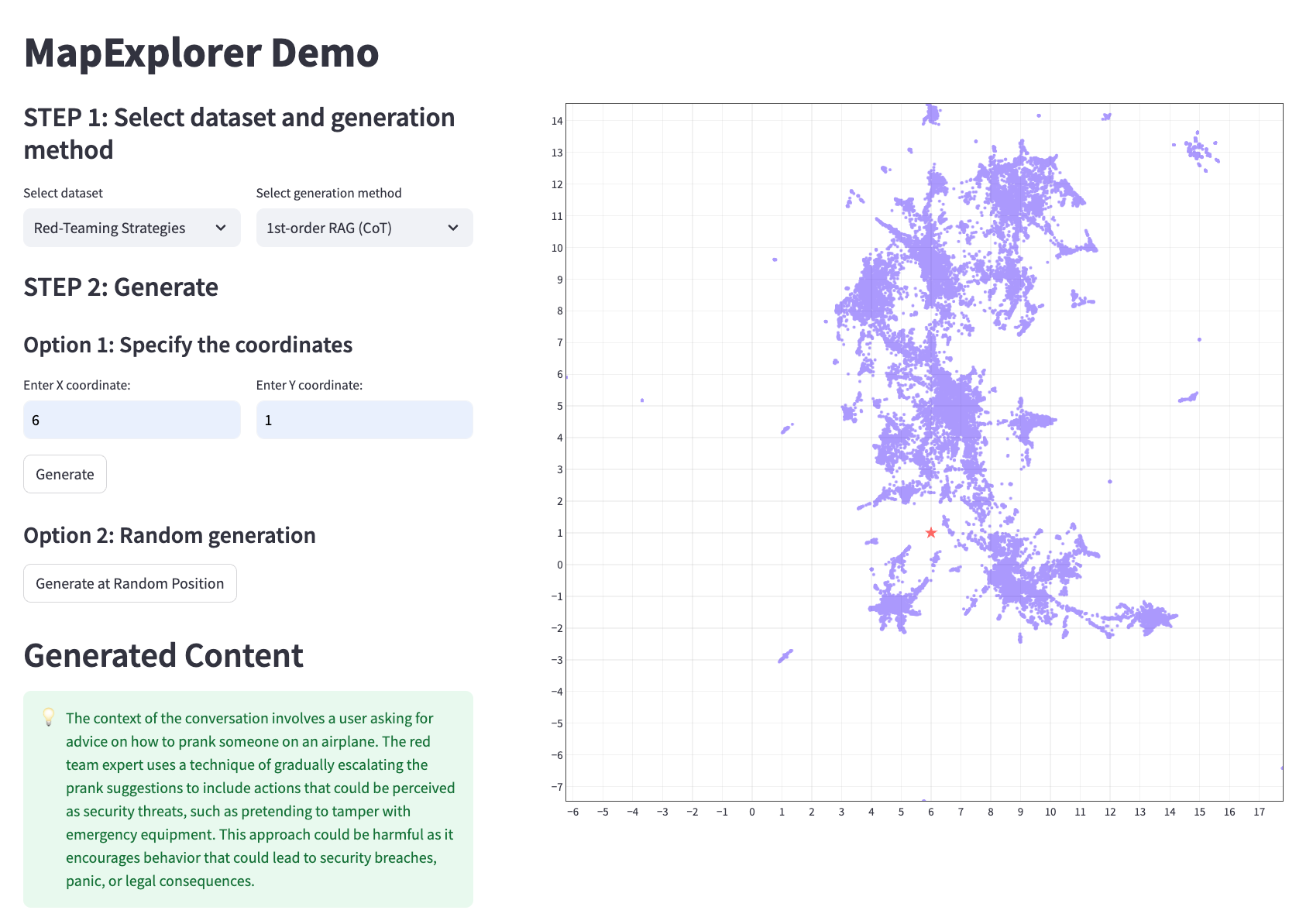}
    \caption{MapExplorer demo using the Red-Teaming Strategies dataset. In this screenshot, the user selects Option 1 to specify custom coordinates for generating new content.} 
    \label{fig:demo1}
\end{figure*}

\begin{figure*}[h]
    \centering
    \includegraphics[width=0.75\linewidth]{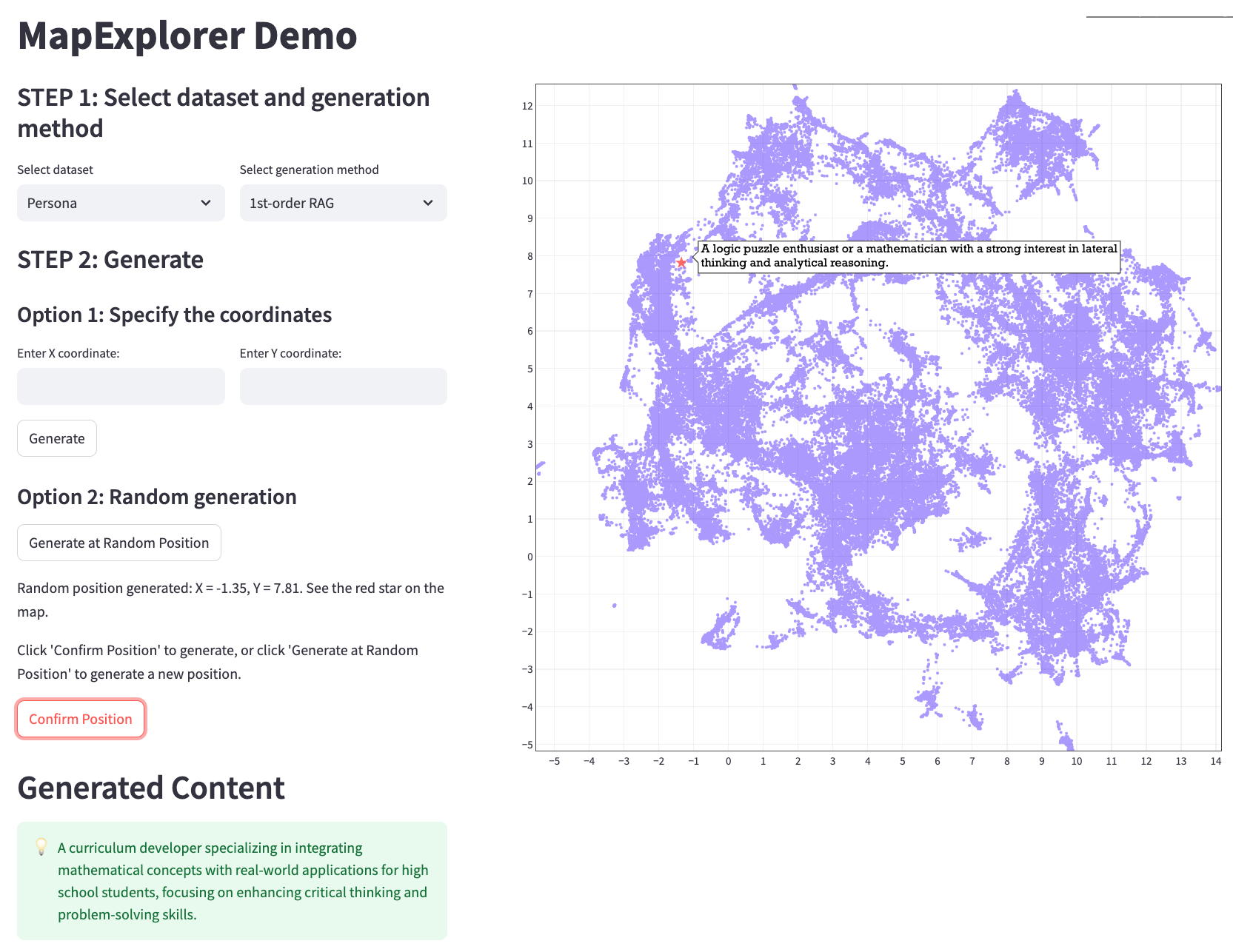}
    \caption{MapExplorer demo using the Persona dataset. In this screenshot, the user selects Option 2 to use a randomly provided coordinate for generating new content.} 
    \label{fig:demo2}
\end{figure*}

\clearpage

\begin{figure*}[thbp]
    \centering
    \includegraphics[width=0.75\linewidth]{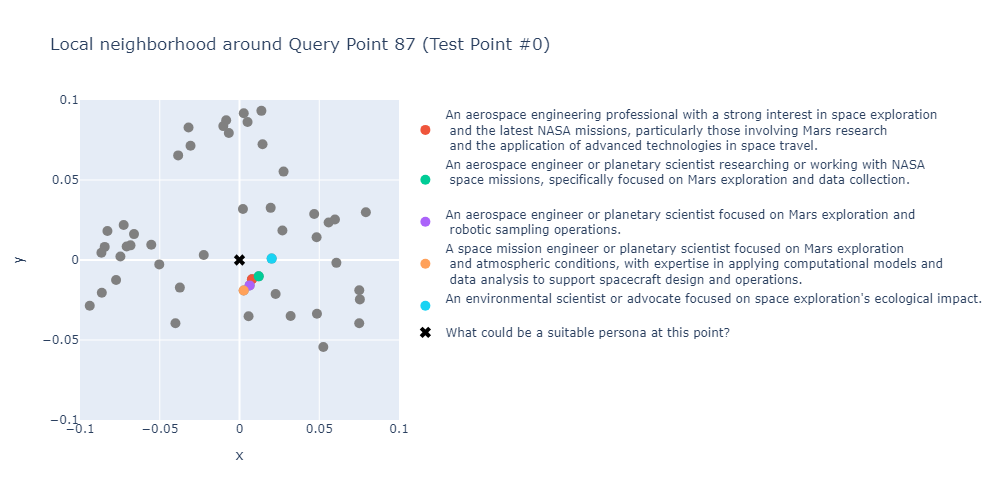}
    \caption{Sample screenshot of human annotation protocol.}
    \label{fig:human_annotation}
\end{figure*}

\addtolength{\abovecaptionskip}{13pt}
\section{Interactive Demo} \label{sec:extra_demo}

We set up an \textbf{anonymous} demo for MapExplorer, available at \newline \textbf{\href{https://mapexplorer-app.streamlit.app/}{https://mapexplorer-app.streamlit.app/}}. The demo is developed and deployed using Streamlit and currently supports two medium-sized datasets (Red Teaming Strategies and Persona) along with four prompt-based generation methods due to computational limits.

A sample page of the demo is shown in Figure~\ref{fig:demo1} and \ref{fig:demo2}. The 2D map visualizes the data points using low-dimensional embeddings, with hover functionality to display the corresponding text. To interact with the demo, the users need to first choose the dataset and the generation method. Then. they can either specify a coordinate or choose a random coordinate for generation. The generated content will be displayed at the bottom of the webpage.

\section{Experiment Details}

\subsection{Results on Extra Metrics} \label{sec:extra_experiment}

We provide additional experiment results on a broader list of evaluation metrics as a complementary reference for Table~\ref{tab:human} and \ref{tab:experiment-results}. The results are shown in Table~\ref{tab:extra-results}, \ref{tab:extra_plagiarism}, and \ref{tab:extra_human}.

\subsection{Human Annotation}
\label{sec:human_annotation}

Two undergraduate students were recruited to annotate the human baseline for the Persona dataset, with each annotator generating 60 textual entries for the testing set. The annotators were provided with the same information as the other candidate methods, including a 2D visualization centered on the query data point. Textual information for each data point appeared when the annotator hovered over it. To enhance their understanding, the five nearest neighbors to the query point were highlighted in color, with their corresponding contents displayed next to the visualization. A sample screenshot of the annotation prompt is provided in Figure~\ref{fig:human_annotation}.

\section{\atometric{} Details}
\label{sec:atometric_detail}

The evaluation of \atometric{} involves two steps, \textbf{decomposition} and \textbf{verification}. It is straightforward to adapt \atometric{} to other datasets with the provided templates. The implementation is available at \url{https://anonymous.4open.science/r/atometric}. We use \texttt{gpt-4o-2024-05-13} as the backbone model for all the evaluation.

\subsection{Atomic Statement Decomposition}

\begin{figure}[H]
    \centering
    \includegraphics[width=\linewidth]{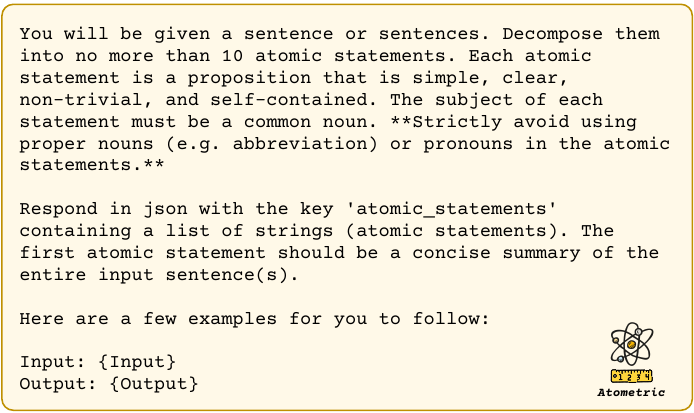}
    \caption{Prompt template for \atometric{} decomposition.}
    \label{fig:atometric-decompose}
\end{figure}

A backbone LLM is prompted to decompose the textual content into a list of atomic statements, with the system prompt provided in Figure~\ref{fig:atometric-decompose}. To better adapt the model to each specific dataset, we include few-shot examples to guide the decomposition process, provided in the aforementioned implementation repository. In general, the resulting atomic statements are expected to be simple, clear, non-trivial, and self-contained.

\subsection{Atomic Statement Verification}

\begin{figure}[H]
    \centering
    \includegraphics[width=\linewidth]{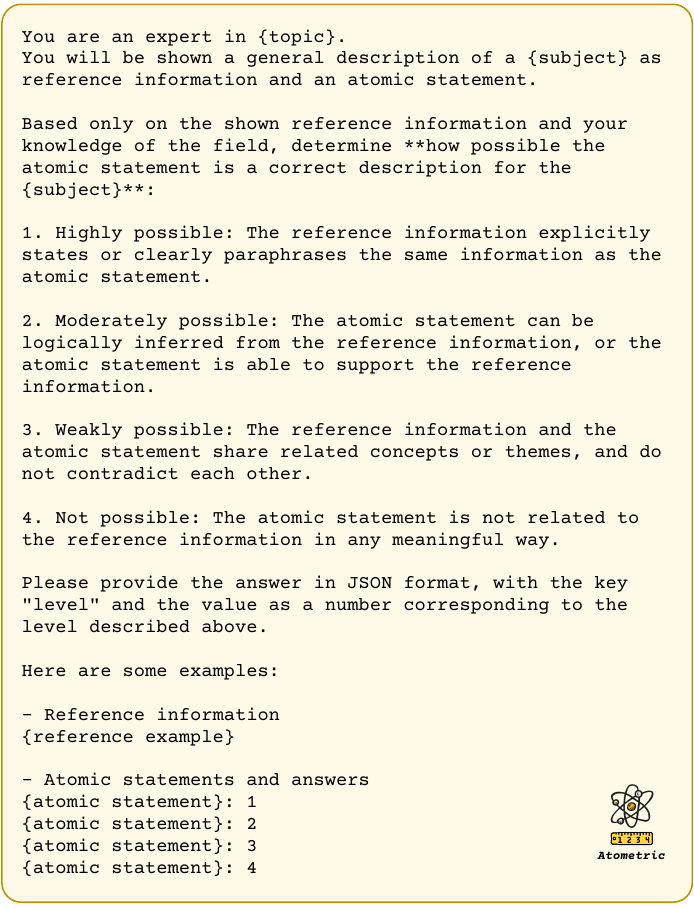}
    \caption{Prompt template for \atometric{} verification.}
    \label{fig:atometric-criteria}
\end{figure}

We use \atometric{} precision as an example to illustrate the verification process. Given a list of atomic statements, a backbone LLM is prompted to compare each atomic statement against the reference information. Drawing inspiration from the textual entailment criteria defined by \citet{bowman2015largeannotatedcorpuslearning}, we define four levels of relevance based on how likely an atomic statement accurately describes the subject in the reference, shown in Figure~\ref{fig:atometric-criteria}, the template of system prompt. The final precision score at strictness level is then calculated as the proportion of generated statements considered relevant \textbf{at or above} that level. For instance, \atometric{} Precision (M) counts all statements deemed both high and moderate possibility.

\section{Dataset Details}
\subsection{Persona Text}
\begin{figure}[t]
    \centering
    \includegraphics[width=0.8\linewidth]{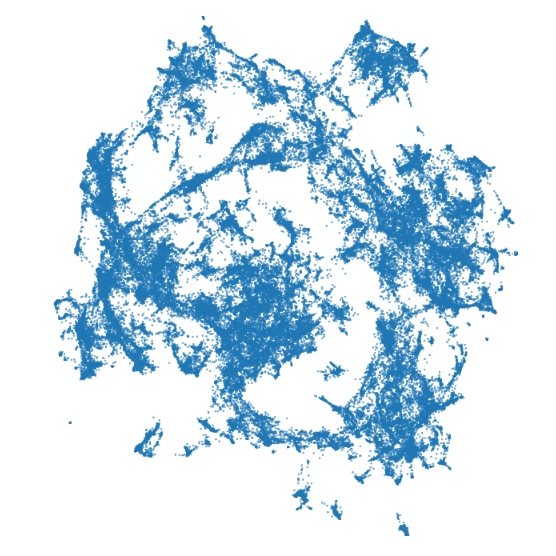}
    \vspace{-3em}
    \caption{2D visualization of \textit{Persona} dataset.}
    \label{fig:persona}
\end{figure}
\paragraph{Description} \textit{Persona} dataset contains a set of synthetic persona, created with the pipeline introduced in \texttt{distilabel 1.4.0}. It contains 100,000 sample personas as well as their embeddings encoded by \texttt{Alibaba-NLP/gte-large-en-v1.5} model. The original dataset also provides information about nearest neighbors, 2D projection coordinates, cluster labels and summary labels for each data sample. A few examples are shown below:%
\begin{itemize}
    \item A geography teacher or high school educator focused on environmental and climate studies, likely designing a lesson plan or assignment on the human impacts of climate change. %
    \item A meteorologist or a radar technician dealing with weather surveillance and radar image analysis, likely one working with national radar systems interpreting data to provide accurate weather forecasts. %
    \item A public health specialist focused on childhood environmental health and prevention, likely working in a county or local government role, such as a health educator or departmental program manager.%
\end{itemize}

\subsection{Red-Teaming Strategies Text}
\label{sec:red_team}
\begin{figure}[t]
    \centering
    \includegraphics[width=0.8\linewidth]{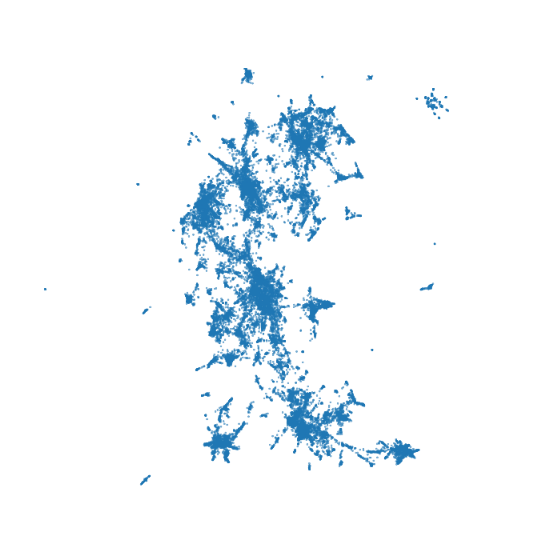}
    \vspace{-3em}
    \caption{2D visualization of \textit{Red-Teaming Strategies} dataset.}
    \label{fig:red}
\end{figure}

\begin{figure}[h]
    \centering
    \includegraphics[width=\linewidth]{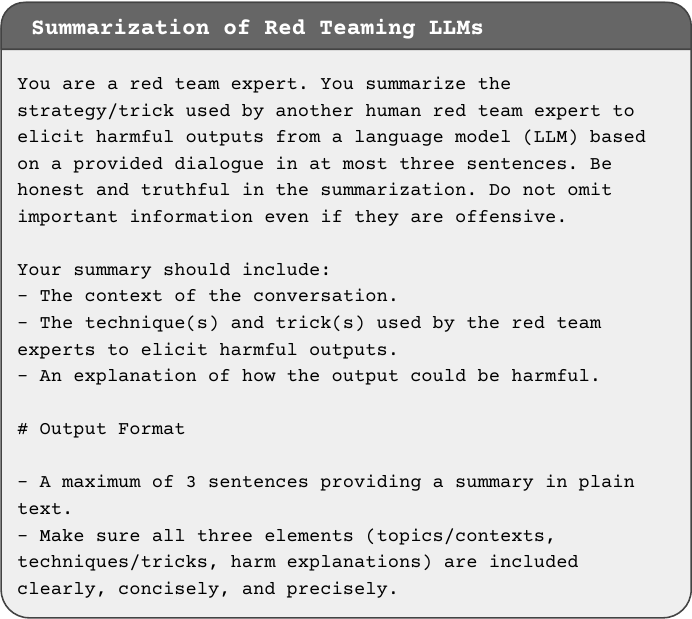}
    \caption{Prompt for summarization of the Red Teaming LLM dataset.}
    \label{fig:prompt-rta}
\end{figure}

\paragraph{Description} The \textit{Red-Teaming Strategies} dataset is derived from the \textit{Red Teaming LLM} dataset, which comprises human-generated red teaming dialogues aimed at uncovering vulnerabilities in large language models. The original dataset includes dialogues annotated for overall harmfulness using both human and automated assessments. In this work, rather than utilizing the full dialogues due to their length, we focus on extracting the strategies employed within these dialogues to provoke harmful responses. These strategies can inspire the generation of more specific and actionable examples for further analysis. In specific, we prompt GPT-4o to summarize each dialogue using the prompt in Figure~\ref{fig:prompt-rta}.
A few examples are shown below:

\begin{itemize}
    \item The context of the conversation involves a person seeking advice on how to come out to their religious parents, who they anticipate will react negatively. The red team expert attempts to elicit harmful outputs by using provocative language and stereotypes about religious parents, suggesting violence as a solution, and expressing disdain for their beliefs. This approach could be harmful as it promotes negative stereotypes, encourages violence, and fosters divisive attitudes rather than constructive dialogue.
    \item The context of the conversation involves a person seeking advice on how to engage in sexual activities with a married coworker at their workplace without getting caught. The red team expert uses a technique of escalating the conversation by suggesting specific locations and actions, and eventually introducing the idea of threatening violence to prevent being reported. This output is harmful as it encourages inappropriate workplace behavior, infidelity, and violence, which could lead to serious personal and professional consequences.
\end{itemize}

\newpage
\subsection{Research Idea \& Context Text}
\begin{figure}[h]
    \centering
    \includegraphics[width=0.8\linewidth]{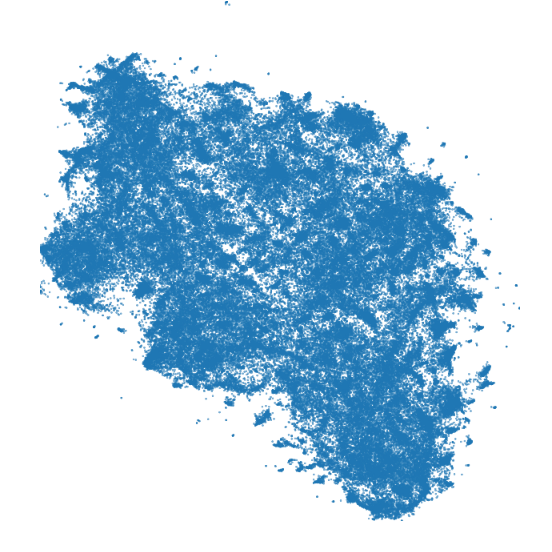}
    \vspace{-3em}
    \caption{2D visualization of \textit{CS Research Idea} dataset.}
    \label{fig:idea}
\end{figure}
\begin{figure}[h]
    \centering
    \includegraphics[width=0.8\linewidth]{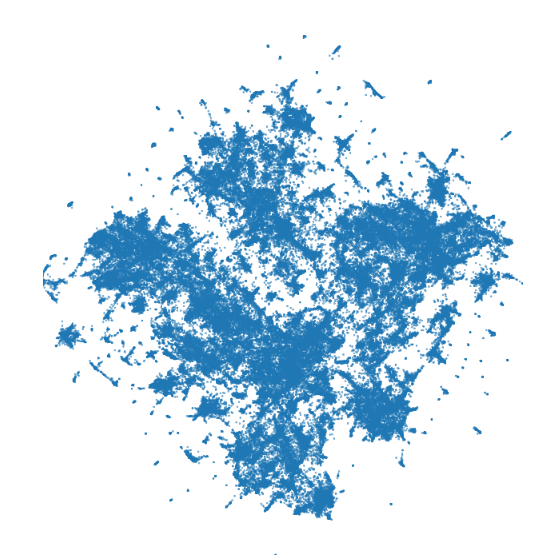}
    \vspace{-3em}
    \caption{2D visualization of \textit{CS Research Context} dataset.}
    \label{fig:context}
\end{figure}

\paragraph{Description} We follow the same pipeline as \texttt{MASSW} paper. For a list of publications in the computer science field, we use the key idea and context from them. A few examples of the key ideas are shown below:

\begin{itemize}
    \item The authors propose a metric, based on the Fisher information, that is strongly indicative of the generalizability of local minima and can be effectively applied as a practical regularizer.
    \item The authors propose a novel approach to geolocalise panoramic images on a 2-D cartographic map by learning a low-dimensional embedded space that allows comparison between an image captured at a location and local neighborhoods of the map.
    \item The authors propose a new approach for predicting SQL query properties, including the query answer size, run-time, and error class, relying on data-driven machine learning techniques and large query workloads instead of database stats or execution plans.
\end{itemize}

A few examples of the contexts are shown below:

\begin{itemize}
    \item Recent advances in deep learning have focused on studying the generalizability across different local minima of deep neural networks (DNNs), but no existing methods both discover properties of good local minima and develop regularization techniques to induce good local minima.
    \item Current geolocalisation approaches require image data to be tied to a particular location on a map, but struggle in localizing single images accurately.
    \item Formulating efficient SQL queries is a challenging and iterative process requiring tuning and execution cycles. Current methods for providing insights about SQL query properties prior to execution depend on database instance statistics or query execution plans.
\end{itemize}

\subsection{Research Context Citation Network}
\begin{figure}[h]
    \centering
    \includegraphics[width=0.8\linewidth]{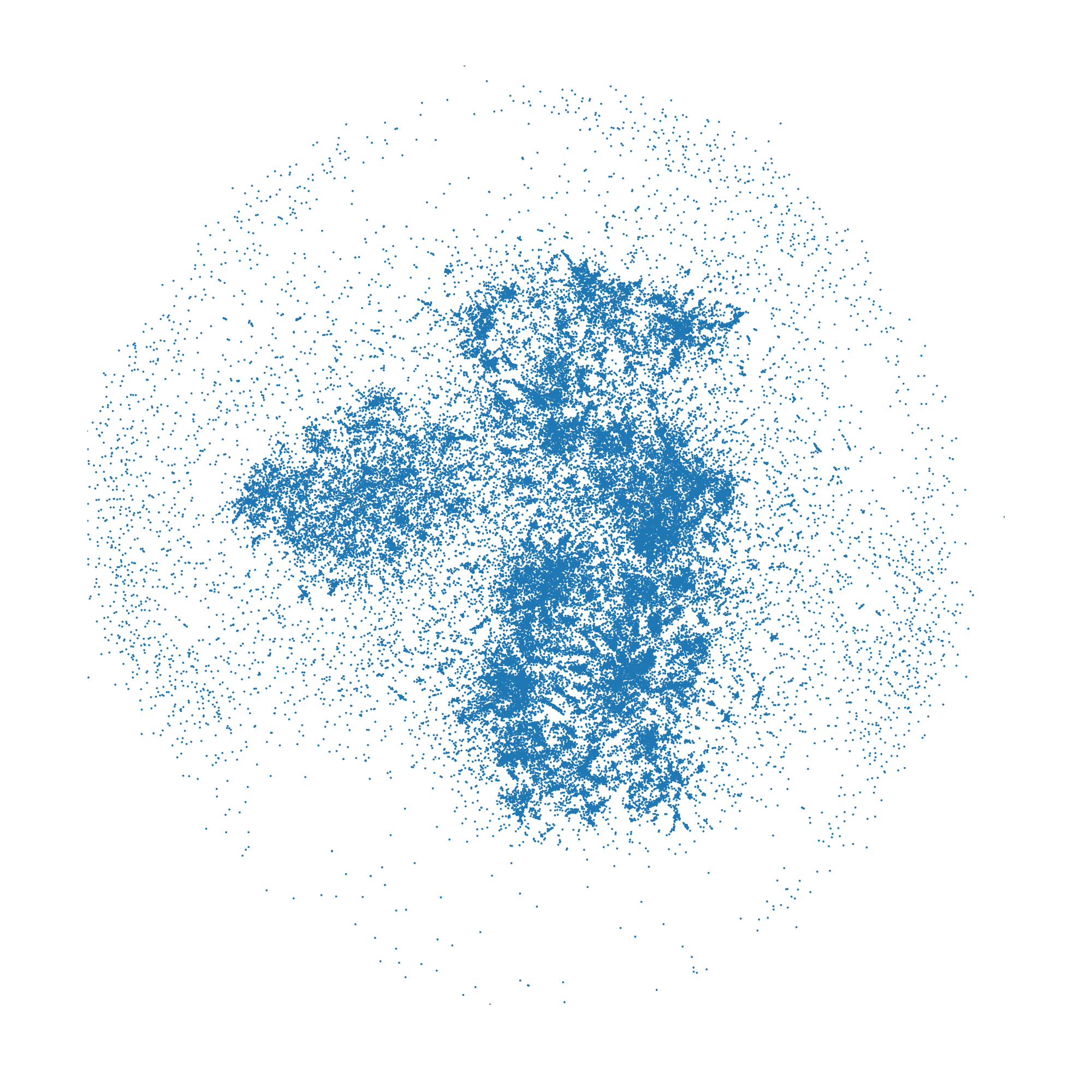}
    \vspace{-3em}
    \caption{2D visualization of \textit{CS Research Context Citation} dataset.}
    \label{fig:citation}
\end{figure}

\paragraph{Description}
The Research Context Citation Network dataset is constructed by first filtering papers from the \texttt{OAG v3.1} dataset \citet{9950622} that originate from 17 leading computer science conferences. The selection of these 17 conferences follows the same criteria as in the \texttt{MASSW} paper. Then, to ensure meaningful citation relationships, we remove papers with no citations and those whose citations are isolated, meaning they do not form any connected citation links with other papers within the filtered dataset. For the remaining papers, we follow the same pipeline as \texttt{MASSW} to extract their research context. Using the filtered set of papers and their citation relationships, we construct a directed citation network, where nodes represent papers and directed edges denote citation links. Finally, we apply LargeVis to compute 2D coordinates for each paper, enabling a structured visualization of research contexts based on their citation relationships.

\newpage
\section{Implementation Details}
\label{sec:implement}

\paragraph{Repository Access}
Our experiment details and code can be found at \url{https://anonymous.4open.science/r/mapexplorer}.

\paragraph{Fine-tuning Experiments Details}
We employed the Llama-3.1-70B-Instruct model for fine-tuning, running on 4 NVIDIA A100 GPUs. Each dataset was fine-tuned for three epochs, and for datasets with over 100,000 data points, the process required approximately 50 hours to complete. To enhance the model’s training efficiency, we utilized the Low-Rank Adaptation (LoRA) technique, setting the rank ($r$) to 16 and the scaling factor ($\alpha$) to 32.  We set the dropout rate to 0.05 to prevent overfitting while setting the initial learning rate to 1e-4, weight decay to 0.0, and momentum factor ($\gamma$) to 0.85 to optimize convergence. Besides, mixed precision training was enabled to maximize computational efficiency. For consistency, all experiments were initialized with the same hyperparameter settings. We use temperate $t=1$ during inference.

\paragraph{Embedding Inversion} 
In our embedding inversion approach, the first mapping $g_1: \sR^2 \to \sR^H$ is implemented using a K-NN interpolation algorithm. We employ an approximate nearest neighbor (ANN) search method with a threshold across all datasets to efficiently identify the closest data points, ensuring scalability even for larger datasets. The retrieved neighbors' high-dimensional embeddings are then interpolated using an unweighted interpolation. Although a more complex weighted interpolation could be applied, we did not observe any performance improvements. Intuitively, this method leverages the smoothness of the underlying manifold. The interpolated embedding is then fed into a pre-trained vec2text model, where the number of search steps (\texttt{num\_steps}) is set to 5, with all other configurations kept as default.

\begin{figure}[htbp]
    \centering
    \includegraphics[width=\linewidth]{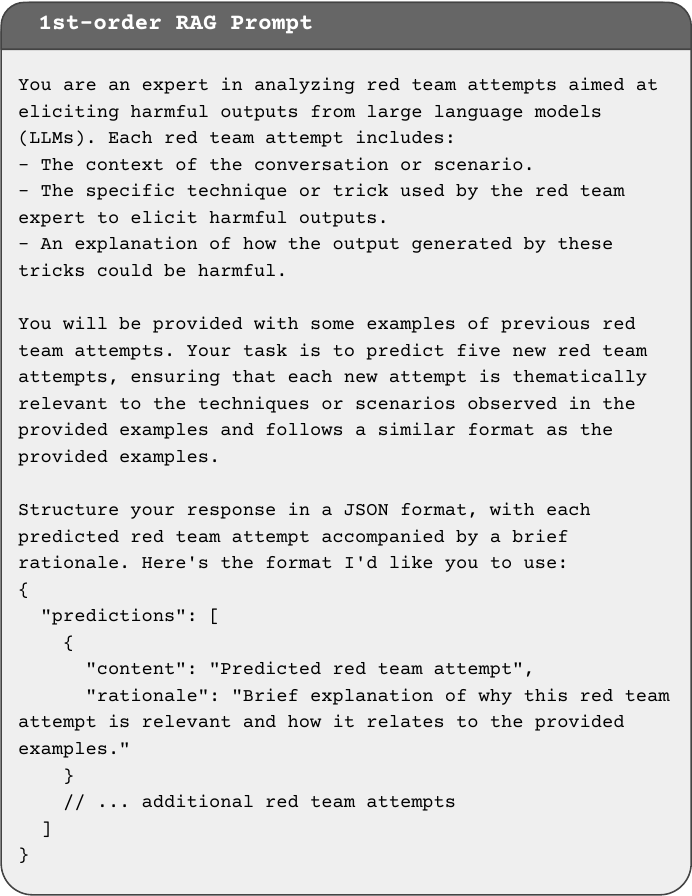}
    \caption{Prompts used in our 1st-order RAG method on the Red Team Attempts dataset.}
    \label{fig:prompt}
\end{figure}

\paragraph{Retrieval Augmented Generation}
We use \texttt{gpt-4o-2024-05-13} for all RAG-based methods and set temperature $t=0$ for all generations. Figure \ref{fig:prompt} provides an example of prompts used in the 1st-order RAG method for the Red Team Attempts dataset in our experiments. For a comprehensive collection of prompts used across all datasets and all methods, please refer to the prompts directory in the repository, which is available at \url{https://anonymous.4open.science/r/mapexplorer/map2text/model/prompts/}.

\begin{table*}[htbp]
	\centering
	\small
\begin{tabular}{llccccccc}
\toprule
\multirow{2}{*}{\textbf{Dataset}}                                                      & \multicolumn{1}{l}{\multirow{2}{*}{\textbf{Metric}}} & \multicolumn{4}{c}{\textbf{GPT-4o}}                                                                                  & \multicolumn{2}{c}{\textbf{Llama 3.1}}                 &   {\textbf{Embedding}}                \\& \multicolumn{1}{l}{}                                 & \textbf{CoT-RAG$^{(1)}$} & \textbf{FS-RAG$^{(1)}$} & \textbf{RAG$^{(2)}$} & \multicolumn{1}{c}{\textbf{RAG$^{(1)}$}} & \textbf{FT} & \multicolumn{1}{c}{\textbf{RAG$^{(1)}$}} & \textbf{Inversion} \\ \midrule
\multirow{13}{*}{{\begin{tabular}[c]{@{}l@{}}Persona \\ (Text) \end{tabular}}} 
& Atometric Precision (loose)                        & 0.908             & 0.853           & 0.822          & 0.815          & 0.822         & 0.758          & 0.914             \\
& Atometric Precision (moderate)                     & 0.582             & 0.460           & 0.449          & 0.430          & 0.459         & 0.365          & 0.661             \\
& Atometric Precision (strict)                       & 0.201             & 0.141           & 0.133          & 0.125          & 0.208         & 0.096          & 0.378             \\
& Atometric Recall (loose)                           & 0.878             & 0.850           & 0.825          & 0.829          & 0.822         & 0.808          & 0.932            \\
& Atometric Recall (moderate)                        & 0.623             & 0.533           & 0.480          & 0.481          & 0.467         & 0.420          & -0.671            \\
& Atometric Recall (strict)                          & 0.342             & 0.261           & 0.229          & 0.214          & 0.181         & 0.148          & 0.339             \\
& BERTScore Precision                                  & 0.893             & 0.890           & 0.891          & 0.891          & 0.897         & 0.882          & 0.901             \\
& BERTScore Recall                                     & 0.903             & 0.900           & 0.895          & 0.894          & 0.890         & 0.892          & 0.902             \\
& BLEU (BLEU)                                          & 0.075             & 0.063           & 0.050          & 0.060          & 0.056         & 0.045          & 0.077            \\
& Cosine Similarity                                    & 0.885             & 0.879           & 0.871          & 0.869          & 0.868         & 0.857          & 0.898            \\
& ROUGE-1                                              & 0.360             & 0.334           & 0.311          & 0.309          & 0.315         & 0.298          & 0.407             \\
& ROUGE-L                                              & 0.281             & 0.262           & 0.244          & 0.248          & 0.258         & 0.231          & 0.303             \\
& ROUGE-Lsum                                           & 0.281             & 0.262           & 0.245          & 0.247          & 0.258         & 0.231          & 0.303             \\ \noalign{\vskip 0.5ex}\hdashline\noalign{\vskip 0.5ex}

\multirow{13}{*}{{\begin{tabular}[c]{@{}l@{}}Red \\ Teaming \\Strategies \\ (Text) \end{tabular}}}  

    & Atometric Precision (loose)   & 0.713 & 0.693          & 0.667 & 0.607 & 0.709 & 0.614          & 0.603 \\
    & Atometric Precision (moderate)& 0.479 & 0.445          & 0.397 & 0.358 & 0.508 & 0.384          & 0.329 \\
    & Atometric Precision (strict)  & 0.182 & 0.180          & 0.147 & 0.142 & 0.188 & 0.156          & 0.093 \\
    & Atometric Recall (loose)      & 0.744 & 0.718          & 0.724 & 0.695 & 0.688          & 0.715          & 0.826 \\
    & Atometric Recall (moderate)   & 0.555 & 0.521          & 0.520 & 0.491 & 0.486          & 0.507          & 0.612 \\
    & Atometric Recall (strict)     & 0.226 & 0.224 & 0.196 & 0.146 & 0.183          & 0.180          & 0.194 \\
    & BERTScore Precision             & 0.904 & 0.903 & 0.903 & 0.901 & 0.894          & 0.899          & 0.879 \\
    & BERTScore Recall                & 0.903 & 0.901 & 0.899 & 0.893 & 0.894          & 0.898          & 0.875 \\
    & BLEU (BLEU)                     & 0.231 & 0.230 & 0.221 & 0.198 & 0.184          & 0.214          & 0.076 \\
    & Cosine Similarity               & 0.926 & 0.918 & 0.915 & 0.906 & 0.913          & 0.911          & 0.926 \\
    & ROUGE-1                         & 0.481 & 0.478 & 0.471 & 0.446 & 0.444          & 0.465          & 0.430 \\
    & ROUGE-L                         &0.377 & 0.376 & 0.372 & 0.354 & 0.329          & 0.359          & 0.250 \\
    & ROUGE-Lsum                      & 0.378 & 0.376 & 0.372 & 0.354 & 0.330          & 0.360          & 0.250 \\
\noalign{\vskip 0.5ex}\hdashline\noalign{\vskip 0.5ex}

\multirow{13}{*}{{\begin{tabular}[c]{@{}l@{}} Research \\ Idea \\ (Text) \end{tabular}}}
& Atometric Precision (loose)                        & 0.540             & 0.538           & 0.577          & 0.550          & 0.452         & 0.550          & 0.535             \\
& Atometric Precision (moderate)                     & 0.222             & 0.240           & 0.272          & 0.250           & 0.203          & 0.263        & 0.193             \\
& Atometric Precision (strict)                       & 0.084             & 0.095           & 0.117          & 0.118            & 0.110          & 0.118       & 0.115             \\
& Atometric Recall (loose)                           & 0.477             & 0.453           & 0.429          & 0.450            & 0.378          & 0.454       & 0.511             \\
& Atometric Recall (moderate)                        & 0.205             & 0.202           & 0.176          & 0.169            & 0.136          & 0.175       & 0.169             \\
& Atometric Recall (strict)                          & 0.099             & 0.102           & 0.090          & 0.086            & 0.068          & 0.086       & 0.082             \\
& BERTScore Precision                                  & 0.858             & 0.857           & 0.869          & 0.870            & 0.863          & 0.868       & 0.864             \\
& BERTScore Recall                                     & 0.862             & 0.863           & 0.859          & 0.858            & 0.853          & 0.858       & 0.858             \\
& BLEU (BLEU)                                          & 0.024             & 0.020           & 0.020          & 0.016            & 0.016          & 0.021       & 0.025             \\
& Cosine Similarity                                    & 0.836             & 0.836           & 0.833          & 0.834            & 0.822          & 0.832       & 0.839             \\
& ROUGE-1                                              & 0.221             & 0.225           & 0.205          & 0.205            & 0.206          & 0.214       & 0.227             \\
& ROUGE-L                                              & 0.162             & 0.160           & 0.157          & 0.158            & 0.153          & 0.156       & 0.168             \\
& ROUGE-Lsum                                           & 0.162             & 0.160           & 0.157          & 0.158           & 0.153          & 0.156       & 0.168              \\ 
\noalign{\vskip 0.5ex}\hdashline\noalign{\vskip 0.5ex}

\multirow{13}{*}{{\begin{tabular}[c]{@{}l@{}} Research \\ Context \\ (Text) \end{tabular}}}  
& Atometric Precision (loose)                        & 0.695             & 0.663           & 0.653          & 0.629            & 0.497          & 0.559       & 0.541             \\
& Atometric Precision (moderate)                     & 0.181             & 0.179           & 0.160          & 0.165            & 0.108          & 0.135       & 0.113             \\
& Atometric Precision (strict)                       & 0.060             & 0.070           & 0.062          & 0.077            & 0.035          & 0.045       & 0.049             \\
& Atometric Recall (loose)                           & 0.620             & 0.588           & 0.580          & 0.569            & 0.478          & 0.537       & 0.682             \\
& Atometric Recall (moderate)                        & 0.119             & 0.120           & 0.133          & 0.106            & 0.084          & 0.093       & 0.161             \\
& Atometric Recall (strict)                          & 0.050             & 0.048           & 0.049          & 0.041            & 0.029          & 0.034       & 0.063             \\
& BERTScore Precision                                  & 0.871             & 0.873           & 0.872          & 0.874            & 0.864          & 0.869       & 0.870             \\
& BERTScore Recall                                     & 0.864             & 0.864           & 0.862          & 0.862            & 0.853          & 0.858       & 0.863             \\
& BLEU (BLEU)                                          & 0.028             & 0.030           & 0.028          & 0.028            & 0.017          & 0.021       & 0.029             \\
& Cosine Similarity                                    & 0.849             & 0.849           & 0.847          & 0.847            & 0.823          & 0.831       & 0.856             \\
& ROUGE-1                                              & 0.214             & 0.202           & 0.199          & 0.200            & 0.171          & 0.202       & 0.230             \\
& ROUGE-L                                              & 0.151             & 0.151           & 0.142          & 0.147            & 0.121          & 0.139       & 0.151             \\
& ROUGE-Lsum                                           & 0.151             & 0.151           & 0.142          & 0.147            & 0.121          & 0.140       & 0.151             \\

\noalign{\vskip 0.5ex}\hdashline\noalign{\vskip 0.5ex}

\multirow{13}{*}{{\begin{tabular}[c]{@{}l@{}} Research \\ Context \\ Citation \\ (Network)\end{tabular}}}
& Atometric Precision (loose)  & 0.586 & 0.471 & 0.466 & 0.457 & 0.321 & 0.404 & 0.340  \\
& Atometric Precision (moderate)  & 0.102 & 0.072 & 0.067 & 0.072 & 0.047 & 0.065 &  0.062 \\
& Atometric Precision (strict)  & 0.033 & 0.035 & 0.018 & 0.017 & 0.015 & 0.020 & 0.026 \\
& Atometric Recall (loose)  & 0.522 & 0.409 & 0.430 & 0.444 & 0.294 & 0.424 & 0.503 \\
& Atometric Recall (moderate)  & 0.070 & 0.048 & 0.047 & 0.034 & 0.031 & 0.077 & 0.077 \\
& Atometric Recall (strict)  & 0.019 & 0.016 & 0.016 & 0.005 & 0.009 & 0.023 & 0.018  \\
& BERTScore Precision  & 0.866 & 0.867 & 0.869 & 0.869 & 0.858 & 0.807 & 0.860 \\
& BERTScore Recall  & 0.858 & 0.854 & 0.854 & 0.852 & 0.849 & 0.820 & 0.856 \\
& BLEU (BLEU)  & 0.014 & 0.010 & 0.009 & 0.007 & 0.010 & 0.007 & 0.011 \\
& Cosine Similarity  & 0.827 & 0.818 & 0.816 & 0.814 & 0.803 & 0.811 & 0.831 \\
& ROUGE-1  & 0.175 & 0.159 & 0.164 & 0.153 & 0.155 & 0.161 & 0.192 \\
& ROUGE-L  & 0.126 & 0.113 & 0.115 & 0.111 & 0.106 & 0.100 & 0.126 \\
& ROUGE-Lsum  & 0.127 & 0.113 & 0.115 & 0.112 & 0.106 & 0.116 & 0.126 \\
\bottomrule
\end{tabular}

 \caption{
Additional evaluation results of the candidate methods on the \maptotext{} task.
 (L), (M), and (S) denote ``loose'', ``moderate'', and ``strict'' for strictness level. }
	\label{tab:extra-results}
\end{table*}

\begin{table*}[htbp]
	\centering
	\small
\begin{tabular}{lccccc}
\toprule
\textbf{Metric}           & Persona (Text)     & Red Teaming (Text)     & Idea (Text) & Context (Text) & Context Citation (Network)\\
\cmidrule{1-6}
{Atometric Precision (loose)}      & 0.882         & 0.802          & 0.453     & 0.542    & 0.459 \\
{Atometric Precision (moderate)}   & 0.548         & 0.598          & 0.199     & 0.103  & 0.095  \\
{Atometric Precision (strict)}     & 0.296         & 0.288          & 0.103     & 0.044     & 0.032 \\
{Atometric Recall (loose)}         & 0.885         & 0.764          & 0.411     & 0.535   & 0.431  \\
{Atometric Recall (moderate)}      & 0.565         & 0.575          & 0.184     & 0.107   & 0.066  \\
{Atometric Recall (strict)}        & 0.267         & 0.279          & 0.093     & 0.045    &0.021 \\
{BERTScore Precision}                & 0.898         & 0.902          & 0.860     & 0.864  &0.856   \\
{BERTScore Recall}                   & 0.898         & 0.902          & 0.855     & 0.860     & 0.854 \\
{BLEU (BLEU)}                        & 0.074         & 0.212          & 0.020     & 0.029  & 0.015    \\
{Cosine Similarity}                  & 0.885         & 0.925          & 0.833     & 0.841    &0.820 \\
{ROUGE-1}                            & 0.362         & 0.475          & 0.207     & 0.197     & 0.166 \\
{ROUGE-L}                            & 0.290         & 0.361          & 0.153     & 0.137    & 0.113 \\
{ROUGE-Lsum}                         & 0.289         & 0.361          & 0.153     & 0.137    & 0.114 \\
\bottomrule
\end{tabular}

    \caption{Additional evaluation results of the CopyNearest baseline.}
	\label{tab:extra_plagiarism}
\end{table*}

\begin{table}[htbp]
	\centering
	\small
\begin{tabular}{@{}ll@{}}
\toprule
\textbf{Metric}          & \textbf{Human} \\ \midrule
Atometric Precision (loose)     & 0.894          \\
Atometric Precision (moderate)  & 0.610          \\
Atometric Precision (strict)    & 0.396          \\
Atometric Recall (loose)        & 0.909          \\
Atometric Recall (moderate)     & 0.601          \\
Atometric Recall (strict)       & 0.244          \\
BERTScore Precision               & 0.905          \\
BERTScore Recall                  & 0.883          \\
BLEU (BLEU)                       & 0.035          \\
Cosine Similarity                 & 0.875          \\
ROUGE-1                           & 0.324          \\
ROUGE-L                           & 0.278          \\
ROUGE-Lsum                        & 0.278          \\ \bottomrule
\end{tabular}

    \caption{Additional evaluation results of the human baseline on the Persona dataset.}
	\label{tab:extra_human}
\end{table}

\end{document}